\documentclass{article}

\usepackage{arxiv}

\usepackage[utf8]{inputenc} 
\usepackage[T1]{fontenc}    
\usepackage{hyperref}       
\usepackage{url}            
\usepackage{booktabs}       
\usepackage{amsfonts}       
\usepackage{nicefrac}       
\usepackage{microtype}      
\usepackage{xcolor}         
\usepackage{subcaption}
\usepackage{graphicx}
\usepackage{color}
\usepackage{multirow}
\usepackage{wrapfig}

\newcommand{\YSR}[1]{{\color{blue} {\bf} YSR: #1}}

\title{
Robustness Analysis of Video-Language Models Against Visual and Language Perturbations
}

\author{Madeline C.~Schiappa \\
    University of Central Florida\\ 
	\texttt{madelineschiappa@knights.ucf.edu} \\
	\And
	    Shruti Vyas \\
  University of Central Florida\\ 
  \texttt{shruti@crcv.ucf.edu} \\
	\And
  Hamid Palangi \\
  Microsoft Research \\
  \texttt{hpalangi@microsoft.com} \\
	\And
	Yogesh S. Rawat\thanks{The authors contributed equally as supervisors to this paper.} \\
  University of Central Florida\\ 
  \texttt{yogesh@crcv.ucf.edu} \\
    \And
  Vibhav Vineet\footnotemark[1] \\
  Microsoft Research \\
  \texttt{vivineet@microsoft.com}
}  

\renewcommand{\headeright}{}
\renewcommand{\undertitle}{\vspace{-5ex}}
\renewcommand{\shorttitle}{Robustness Analysis of Video-Language Models Against Visual and Language Perturbations}

\hypersetup{
pdftitle={Robustness Analysis of Video-Language Models Against Visual and Language Perturbations},
pdfsubject={cs.CV},
pdfauthor={Madeline C.~Schiappa, Shruti Vyas, Hamid Palangi, Yogesh S.~Rawat, Vibhav Vineet},
pdfkeywords={robustness, video robustness, text robustness, text-to-video retreival, video retreival, video-language modeling, computer vision},
}

\begin{document}

\maketitle

\begin{abstract}
 Joint visual and language modeling on large-scale datasets has recently shown good progress in multi-modal tasks when compared to single modal learning. 
 However, robustness of these 
 approaches against real-world perturbations has not been studied. In this work, we perform the first extensive robustness study of video-language models against various real-world perturbations. 
 We focus on text-to-video retrieval 
 and propose two large-scale benchmark datasets, \textit{MSRVTT-P} and \textit{YouCook2-P}, which utilize 90 different visual and 35 different text perturbations. The study reveals some interesting initial findings from the studied models: 1) models are generally more susceptible when only video is perturbed as opposed to when only text is perturbed, 2) models that are pre-trained are more robust than those trained from scratch, 3) models attend more to scene and objects rather than motion and action, 4) models showed gender biases when trained on a target dataset as opposed to a larger, pre-training dataset. 
We hope this study will serve as a benchmark and guide future research in robust video-language learning.
The benchmark introduced in this study along with the code and datasets is available at \url{https://bit.ly/3CNOly4}.

  
\end{abstract}

\section{Introduction}
    
Human beings learn different skills sequentially and in a continual manner. 
Sequential data like video and language are natural forms of input to any intelligent vision system operating in the real world. 
Robustness of these intelligent systems against real-world distribution shifts is crucial for various applications including autonomous driving \cite{ma2018improved, guo2019safe, deng2021deep, nesti2022evaluating}, medicine \cite{ardulov2021robust, ali2021deep,itzkovich2019using, sung2020artificial}, robotics \cite{itzkovich2019using, yim2007modular, lakomkin2018robustness, bednarek2020robustness} and others. In a multimodal setting where both language and video are used, these distribution shifts can occur for a variety of reasons. In video, these can include lighting, camera movement, digital compression, etc. In text, these can include spelling errors, incorrect synonym swapping, bias, etc. These distribution shifts can cause deep learning models to fail when deployed 
in a real world setting \cite{hendrycks2018benchmarking,aifail2020soccer, deng2021deep}. 

It is crucial that these models are robust against such distribution shifts for successful deployment. 
Robustness has been an active topic of research in deep learning. However, most of the effort is directed towards robustness against adversarial attacks \cite{chakraborty2021survey,alshemali2020improving, dong2021should}. 
There are some recent efforts on robustness against real-world distribution shifts, but they focus on non-sequential image data \cite{hendrycks2018benchmarking,bhojanapalli2021understanding,hendrycks2021many} and natural language \cite{textflint} independently. 
Because video and text are vital sequential inputs for real-world intelligent systems, studying robustness in a multimodal setting is an important step towards developing reliable systems and has never been studied before.

In this work, we perform a large-scale analysis on the robustness of existing multimodal deep learning models for text-to-video retrieval. Text-to-video retrieval provides an important test scenario for 
a multimodal setting as it evaluates the similarity between video and text embeddings and how their joint-embedding space may vary based on distribution shifts on one or both modalities. 
There are several questions about existing methods which are unanswered.
Are these approaches robust to real-world corruptions in one modality and even both? Do we really need a heavy pre-training strategy for robustness or is training on the target dataset enough? Are the recently introduced transformer-based models better for robustness? Do these approaches utilize temporal modeling? Are these models biased? This study aims to be the first to answer some of these critical questions for video-language deep learning models. 

Towards this goal, we present two benchmark datasets to conduct robustness analysis on text-to-video retrieval. We utilize two widely used retrieval datasets MSRVTT \cite{msrvtt} and YouCook2 \cite{youcook2} and propose corresponding benchmark datasets, \textit{MSRVTT-P} and \textit{YouCook2-P}. In order to create these benchmarks, we introduce 90 different visual perturbations and 35 textual perturbations.

This study reveals several interesting observations about robustness of video-language models: 
1) The studied models are more susceptible when only video is perturbed as opposed to when only text is perturbed. 2) Model pre-training improves both robustness and performance. 3) Models attend more to object and scene rather than motion and action. 4) Models showed gender biases when trained on a target dataset as opposed to a larger, pre-training dataset, even when fine-tuned.

We make the following contributions in this study, 
\begin{itemize}
    \setlength\itemsep{0.05em}
    \item We focus on robustness of video-language approaches against distribution shifts due to spatial/spatio-temporal visual and text perturbations; this problem has not been studied before to the best of our knowledge.
    \item We provide two large-scale benchmark datasets (MSRVTT-P and YouCook2-P) to conduct robustness analysis on text-to-video retrieval. 
    \item We present an empirical analysis of video-language approaches to study the effect of various perturbations on their performance.
\end{itemize}

\begin{figure*}
    \centering
    \includegraphics[width=.80\linewidth]{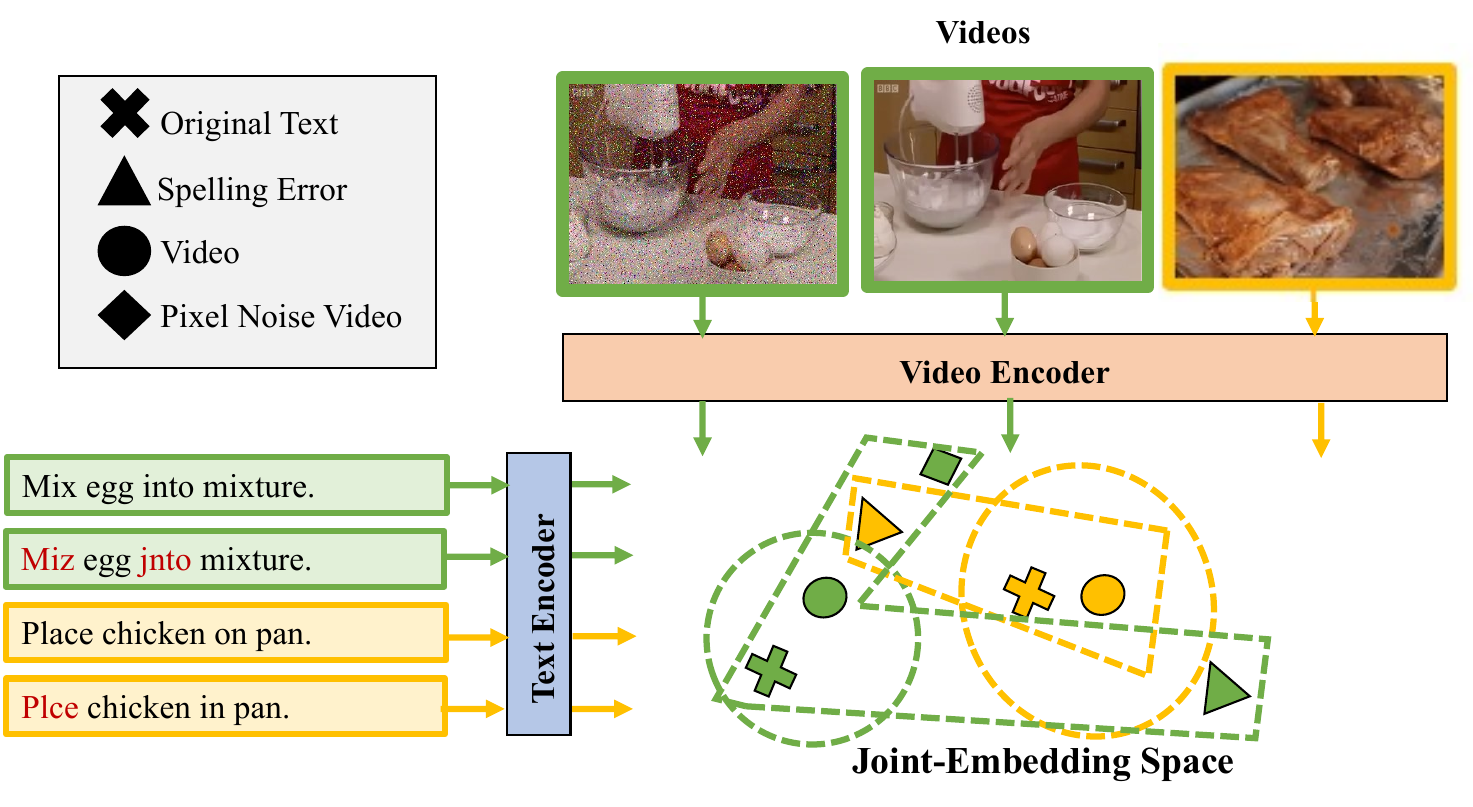}
    \caption{\footnotesize{A conceptual diagram of video and text in a joint latent space where the original text (cross) are closer to their paired video compared to text that is perturbed via typos (triangle) and original videos (circles) are closer to their paired text compared to videos that are perturbed (diamond). Models are considered robust when the perturbed text is still closest to its respective video when the semantic meaning of text is preserved. The same should be true if video is perturbed or both are perturbed.}}
    \label{fig:teaser}
\end{figure*}

\section{Related Works}
\subsection{Robustness}
\paragraph{Visual} 
Most recent works on robustness in the visual domain 
have focused on real-world distribution shifts as opposed to targeted attacks in the image domain \cite{hendrycks2018benchmarking,bhojanapalli2021understanding,hendrycks2021many,sakaridis2021acdc, taori2019robustness}. In \cite{hendrycks2018benchmarking, recht2019imagenet}, authors analyze different image classification models on naturally occurring distribution shifts using ImageNet. While the benchmark study analyzing naturally occurring shifts in \cite{taori2019robustness} demonstrated that data augmentation is not sufficient for robustness, several studies have found that certain data augmentations do improve the robustness of deep learning image models \cite{geirhos2018imagenet,hendrycks2019augmix,yin2019fourier}. These data augmentations are often noise related \cite{madry2018towards,rusak2020increasing,lopes2019improving} but other transformations such as color or texture have been analyzed as well \cite{geirhos2018imagenet,yun2019cutmix,cubuk2019autoaugment,hendrycks2019augmix}. 
These studies have not yet been extended to the video domain where temporal aspects are also present.  Different from these works, this study will provide a benchmark on robustness of models against real-world perturbations in multi-modal settings.

\paragraph{Text}
Research on robustness in the natural-language processing (NLP) field is far more extensive as compared to video. 
Some works in natural distribution shifts focus on semantic changing of a phrase \cite{gardner2020evaluating, schlegel2020semantics}. In \cite{gardner2020evaluating}, the phrase is altered in small, meaningful ways that change the overall label in order to understand the decision boundaries of models. Similarly, \cite{schlegel2020semantics} alter text in ways that change the semantic meaning but keep the original text's lexical surface form. Other works inspired by \cite{hendrycks2021natural} focus on distribution shift based on changes of grammar errors, dialects, speakers, and language \cite{demszky2020learning}, different domains \cite{miller2020effect} and bias \cite{de2019bias, prates2020assessing}. 
The image robustness research space has inspired many of these studies, but there are vast differences to NLP and vision that make these transfers difficult, such as the discrete vs. continuous search space as explained in \cite{wang2021measure}. Data augmentation has also been looked at as a method to improve robustness and has shown substantial improvements \cite{feng2021survey, dhole2021nl, chen2020mixtext, hendrycks2019augmix, chen2021hiddencut}. These studies have not yet been extended to the multimodal domain where vision is also incorporated.  Different from these works, this work will provide a large-scale benchmark on robustness for multimodal models against real-world perturbations.

\paragraph{Multimodal}
Evaluating robustness in multimodal models is more difficult because there are more vectors of attack possible. It is possible to attack the entire model while perturbing only one of the modalities used or a varying amount of the modalities used. Focus on single-modality attacks in \cite{yang2021defending} investigated the robustness of multimodal neural networks against worst-case (i.e., adversarial) perturbations on a single modality. 
Looking at multimodal attacks, \cite{tian2021can} evaluated audio-visual models by running adversarial attacks on audio, visual, and both modalities. A more general benchmark was proposed in \cite{liang2021multibench} for a variety of modalities. Different to this benchmark, we focus on robustness analysis of the video and text embedding space in great detail.
Such studies have not been performed on naturally occurring distribution shifts and have not looked at video-language models specifically, two modalities that are drastically different. 
\subsection{Video-Language Modelling}
Multimodal modeling with text and vision has improved since the emergence of both the HowTo100M dataset \cite{howto100m} and transformer architecture \cite{BERT}. The highest performing models \cite{univl, videoclip, vlm, avlnet, vatt, patricketal2020} pre-train on the \cite{howto100m} and most use pre-extracted visual features from the original multimodal model from \cite{miech2020end} which uses an S3D-G 
backbone \cite{xie2018rethinking}. For learning a joint visual-text space, these models often use a contrastive learning objective between visual and text embeddings \cite{miech2020end, videoclip, vatt, avlnet} while some use an alignment-based objective \cite{univl, vlm} using masked modeling. Many of the contrastive approaches \cite{videoclip, miech2020end, avlnet}, use a two-branch encoder approach where video has one encoder and text a  separate encoder and the objective is to move the two enoder outputs closer to each other in latent space. Some approaches \cite{univl, vlm, vatt} will additionally utilize a cross-encoder before comparing output. This work will provide a greater understanding of these video-language models and their robustness.

\section{Distribution Shift}
Existing research in multimodal learning is mostly focused on training and testing the proposed methods on a benchmark dataset with little to no distribution shift from training to testing samples. While models often use a video encoder that is pre-trained on a very large, noisy dataset, e.g. HowTo100M \cite{howto100m}, there is no understanding of how, in a multimodal setting, a distribution shift will affect the joint-embedding space of video and text. 
To study the effect of distribution shift, we introduce five categories of visual perturbations and seven categories of text perturbations. 
More details about these perturbations are provided in the Appendix.

\subsection{Visual Perturbations}
First, we extend image-based perturbations from \cite{hendrycks2018benchmarking} to videos. Next, we add temporal perturbations to address the time dimension and video compression to address video-specific distribution shifts as well as spatio-temporal. The total set of visual perturbations fall into 5 categories: \textbf{Noise}, \textbf{Blur}, \textbf{Temporal}, \textbf{Camera} and \textbf{Digital}. Each visual perturbation has a severity range from 1 to 5 where the greater the severity, the more challenging and perturbed the video is. Blur, Noise, and Camera perturbations are all applied frame-by-frame. Noise includes \textit{Impulse}, \textit{Gaussian}, \textit{Shot}, and \textit{Speckle}, Blur includes \textit{Zoom}, \textit{Defocus} and \textit{Motion} and Camera includes \textit{StaticRotate}, \textit{Rotation} and \textit{Translation}. 

The Digital and Temporal perturbations are added in order to include distribution shifts specific to video while also perturbing spatially and temporally. \textbf{Digital} perturbations are related to compression and video-streaming quality. We evaluted models on \textit{JPEG}, \textit{MPEG1} and \textit{MPEG2}. JPEG is a lossy image compression, MPEG1 compresses video without excessive quality loss and MPEG2 is a lossy compression for video that is similar to MPEG1. \textbf{Temporal} perturbations focus on the time dimension in a video and include \textit{Sampling}, \textit{Reverse Sampling}, \textit{Jumbling}, \textit{Box Jumbling} and \textit{Freeze} and will help in understanding how these models are utilizing temporal information. Sampling rate slows the playback speed by sampling frames uniformly at a varying level of rates and reverse samping does so in the reverse order of the original sequence. Jumbling splits a video into segments and randomly shuffles the frames in that segment while Box jumbling randomly shuffles the segments. Freezing simulates when live streaming buffers, freezing on random frames for random durations. 

\subsection{Text Perturbations}
\label{sec:text_perturbations}
We group text perturbations into three different types, natural, machine-based, and synthetic. Here machine-based perturbations use a model to alter the text while natural-based imitates real-world mistakes when generating text. Synthetic are not natural but are used to gain a greater understanding of the models. 
The text perturbations are further grouped into seven different categories \textit{ChangeChar}, \textit{AddText}, \textit{Bias}, \textit{Positional}, \textit{DropText}, \textit{SwapText} and \textit{TextStyle} with a total of 35 different perturbations.
\textbf{ChangeChar} refers to any perturbation that changes a character in word(s). \textbf{SwapText} is a machine-learning based perturbation that swaps word(s) from the original phrase. \textbf{AddText} includes appending irrelevant phrases to text or inserting adverbs. 
\textbf{TextStyle} are perturbations that change the original text's style, e.g. making it \textit{passive} \cite{styleformer}. 
\textbf{Bias} perturbations include switching the gender of word(s) in a phrase \cite{genderbender}. We additionally include changing all male references to female, the reverse, and convert all gender-specific references to gender neutral. 

\textbf{DropText} perturbations are synthetic and drop words based on their part-of-speech (POS) tag. 
These perturbations are included to gain a better understanding of word level attention, more specifically, to understand if models attend more to objects, actions or context. \textit{DropNN}, \textit{DropVB}, and \textit{DropVBNN} are different variations of dropping words based on whether the POS tags are Noun and/or Verb. Because there are often more nouns in a sentence, we have an additional perturbation \textit{RandNN} where only one noun is dropped randomly as opposed to all.
For example, ``a little girl does gymnastics'' becomes `` a little [UNK] does gymnastics''. 
In order to evaluate attention to contextual words, \textit{OnlyNN}, \textit{OnlyVB}, and \textit{OnlyNNVB} drops all words but those with POS NN and/or VB. \textbf{Positional} perturbations are machine-based and alter the phrase based off their location. This is used to evaluate the models based on the position of words in a phrase.
These include \textit{DropFirst}, \textit{DropLast}, \textit{DropFirstandLast}, and \textit{ShuffleOrder}. Drop-related perturbations will replace a word at that position with an [UNK] tag. The ShuffleOrder perturbation shuffles the words in a phrase randomly. More details on the generated text perturbation are provided in the Appendix. 

\begin{table}[t!]
    \centering
    \small
    \caption{Details of self-supervised video-language models used in this study.}
    \begin{tabular}{l|c|c|c|c|c}
    \hline 
       Model & Params  & Text Input & Text Encoder & Video Input & Video Encoder \\
       \hline \hline
        HowTo100M MIL \cite{miech2020end} & 31.2M & Raw & Word2Vec \cite{word2vec} & Raw & S3Dg \cite{xie2018rethinking} \\ 
        VideoClip \cite{videoclip} & 177.4M & Raw & BERT \cite{BERT} & S3D \cite{miech2020end} & MLP+Transformer \\ 
        UniVL \cite{univl} & 153.7M & Raw & BERT \cite{BERT} & S3D \cite{miech2020end} & Transformer \\ 
        COOT \cite{COOT} & 7.6M & BERT \cite{BERT} & Transformer & S3D \cite{miech2020end} & Transformer \\ 
        FIT \cite{fit} & 180.9M & Raw & BERT \cite{BERT} & Raw & ViT \cite{vit,timesformer} \\
        \hline
    \end{tabular}
    \label{tab:models}
\end{table}


\section{Robustness Benchmarks and Evaluation}

\subsection{Model Variants}
We perform our experiments on five different self-supervised video-language models which are based on CNN and Transformer architectures. The goal is to benchmark multiple pre-training approaches while simultaneously study the behavior of CNN and transformer based models for robustness in text-to-video retrieval. Models were chosen based on whether they provided 1) a usable code base, 2) model weights, 3) and used text and video as their modalities.

We evaluate the most popular video-language approach MIL-NCE \cite{miech2020end} which uses a CNN backbone and Word2Vec word embeddings with an MIL-NCE contrastive loss between text-video pairs. We further evaluate models and approaches that utilize visual features from \cite{miech2020end} with further training and different self-supervised approaches. The more recent method VideoClip \cite{videoclip} is a transformer-based approach relying instead on BERT \cite{BERT} for both text and video encodings with a similar but improved contrastive loss. COOT \cite{COOT} similarly uses transformer-based encoders taking BERT text features and S3D visual features as input and includes cross-attention between the text and video features. Rather than a contrastive loss with negative pairing, COOT focuses on alignment between text and video alone. UniVL \cite{univl}, is another transformer-based approach that uses a cross-encoding transformer in addition to separate encoders as their self-supervised objective. The final approach evaluated, FIT \cite{fit}, combines image-based research with video. It uses only a small set of frames for a given clip which is encoded using a Visual Transformer (ViT) \cite{vit, timesformer}. They also pre-train with a different dataset that comprises of both images from CC3M \cite{CC3M} and video from their own proposed dataset, Web2Vid \cite{fit}. FIT uses a contrastive loss for video-text pairs and for text-video pairs with temporal curriculum learning. More details on these approaches are shown in Table \ref{tab:models}.

\subsection{Datasets}
We use two video-language datasets for our experiments: MSRVTT \cite{msrvtt} and YouCook2 \cite{youcook2}. \textbf{MSRVTT} is a video captioning dataset which consists of 10,000 clips with an average length of 10 seconds each. These videos show a variety of activities that can be organized into 20 categories. We follow JSFusion \cite{yu2018joint, miech2020end, videoclip} which randomly samples 1K clip-text pairs as test data for evaluation. \textbf{YouCook2} is a task-oriented cooking dataset with 2000 long untrimmed videos from 89 cooking recipes. Each video is annotated with captions with provided temporal boundaries, allowing each video to be split into a set of clips. There are 3,305 test clip-text pairs from 457 videos for evaluation.

Captions in the MSRVTT and YouCook2 dataset are quite different. 
%
YouCook2 has no indication of gender with phrases comprising 2x more nouns compared to MSRVTT while MSRVTT has a more uniform distribution of words with an increased vocab size of 568 more unique words. Videos in MSRVTT and YouCook2 are also different where YouCook2 are long, complex activities split into clips with temporally bounded annotations. The test dataset will have multiple clips from the same video while all test clips in MSRVTT are from different videos. This may make clips less distinguishable when compared to clips from MSRVTT. \textit{ This means the distributions between the two datasets are different and may result in different observations. }


%
We apply 90 visual perturbation to the test videos, 31 or 35 text perturbations to the captions, and 66 visual and text combined perturbations for creating robustness benchmarks \textbf{YouCook2-P} and \textbf{MSRVTT-P}. YouCook2-P does not have gender-related perturbations because of no reference to gender in their captioning, therefore only 31 text perturbations are used. MSRVTT-P consists of 90,000 videos and 35,000 captions resulting in 2,766,000 video-text pairs. YouCook2-P consists of 41,130 videos split into 301,500 clips and 103,850 captions, resulting in 9,266,100 clip-text pairs. 

\subsection{Tasks and Evaluation Metrics}
We evaluate the performance of models on text-to-video retrieval using a retrieval rate R@K metric \cite{miech2020end}. 
To measure robustness, we 
use two metrics; one for absolute retrieval drop and the other for relative retrieval drop \cite{hendrycks2018benchmarking,su2018robustness,taori2020measuring,liang2021multibench}. 
Given a trained classifier model $f$, we first compute retrieval $R_c^{f}$ on the clean test set. Next, we test this classifier on a perturbation $p$ and obtain retrieval $R^f_{p}$ for perturbation $p$. The absolute robustness $\gamma^a$ is computed for each perturbation $p$ as $\gamma^a_{p} = 1- (R^f_c - R^f_{p})/100$. For visual perturbations, the aggregated performance of a model can be obtained by averaging all severity levels to get $\gamma^a_{p}$ and over all perturbations to get $\gamma^a \pm \sigma$. For text perturbations, the aggregated performance of a model can be obtained by averaging across sub-types rather than severity. To take into account differing model performance on the clean dataset, we compute relative performance drop to measure models robustness. The relative robustness $\gamma^r$ is computed for each perturbation $p$ as $\gamma^r_{p} = 1 - (R^f_c - R^f_{p})/R^f_c$ which is the difference normalized to the accuracy of the model on the test set without a perturbation. The robustness score will usually range from 0 to 1, where 0 indicates a model is not robust and 1 is where the model is entirely robust. A score greater than 1 indicates that the model's performance is better with the perturbation. 

\begin{figure}[t!]
    \centering
    \includegraphics[width=.85\textwidth]{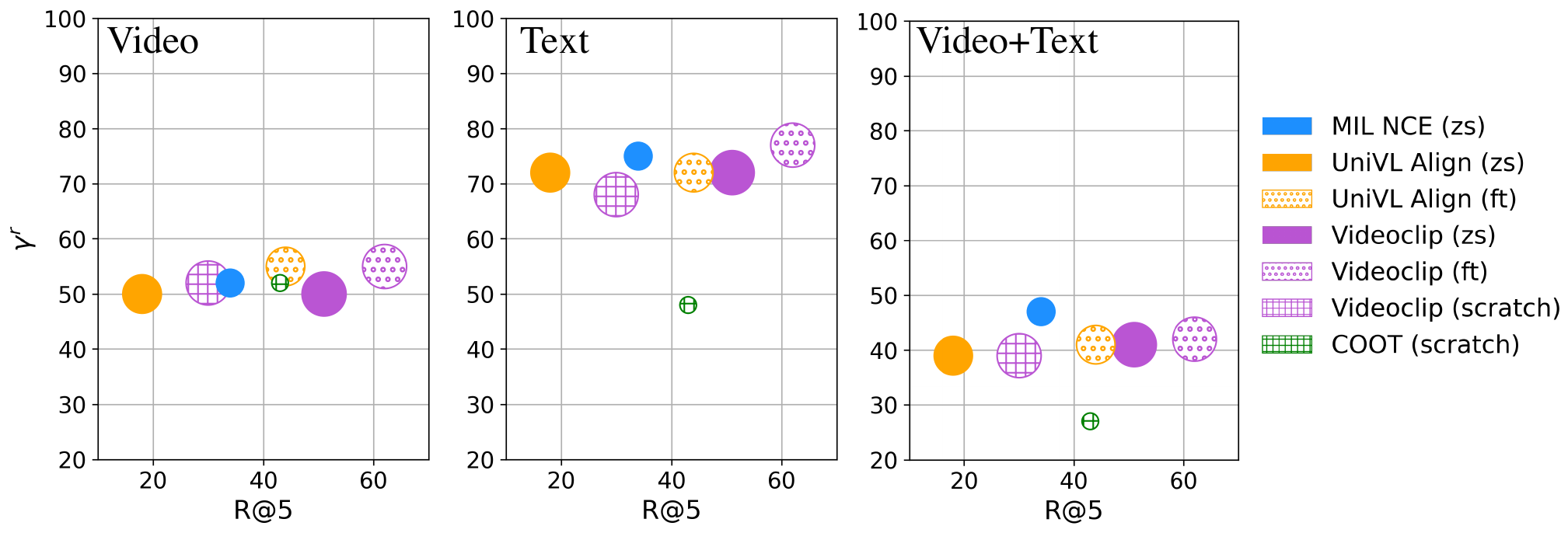}
    \caption{
    A comparison of models under different training protocols (zs: zero-shot learning, ft: finetuning on target dataset, and scratch: no pretraining) on thr YouCook3 dataset. The y-axis is the drop in performance when data is perturbed measured by the relative robustness $\gamma^r$ score, the x-axis is R@5 for text-to-video retrieval, and the size of marker represents number of model parameters. The three plots (left to right) corresponds to visual, text, and visual+text perturbations respectively.
    }
    \label{fig:teaser_yc2}
\end{figure}

\begin{table}[t!]
    \centering
    \caption{The drop in performance when data is perturbed measured by the Relative robustness $\gamma^r$ for each category of video perturbations on MSRVTT-P with SD $\pm \sigma$. The ViT based approach FIT is noticeably more robust on spatial noise as compared to the other approaches evaluated.}
    \label{tab:visual_robustness_scores_msrvtt}
    \resizebox{0.8\textwidth}{!}{
\begin{tabular}{llllll}
\toprule
\textbf{Method} &                      \textbf{ Blur }&                     \textbf{Camera} &                    \textbf{Digital} &                      \textbf{Noise} &                   \textbf{Temporal} \\
\midrule
FIT (scratch)       &              0.67$\pm$0.13 &              0.90$\pm$0.12 &  \underline{0.84$\pm$0.07} &              0.73$\pm$0.24 &  \underline{1.00$\pm$0.01} \\
VideoClip (scratch) &              0.54$\pm$0.23 &              0.80$\pm$0.17 &              0.53$\pm$0.21 &              0.24$\pm$0.20 &              0.96$\pm$0.05 \\
FIT (zs)            &     \textbf{0.79$\pm$0.13} &     \textbf{0.97$\pm$0.10} &     \textbf{0.88$\pm$0.07} &     \textbf{0.81$\pm$0.21} &     \textbf{1.03$\pm$0.02} \\
MIL NCE (zs)        &              0.59$\pm$0.17 &              0.78$\pm$0.11 &              0.42$\pm$0.20 &              0.24$\pm$0.19 &              0.88$\pm$0.06 \\
UniVL (zs)          &              0.61$\pm$0.21 &              0.85$\pm$0.12 &              0.61$\pm$0.16 &              0.27$\pm$0.20 &              0.96$\pm$0.04 \\
VideoClip (zs)      &              0.61$\pm$0.22 &              0.84$\pm$0.13 &              0.62$\pm$0.18 &              0.22$\pm$0.17 &              0.95$\pm$0.02 \\
FIT (ft)            &  \underline{0.74$\pm$0.11} &  \underline{0.92$\pm$0.11} &              0.83$\pm$0.07 &  \underline{0.77$\pm$0.20} &  \underline{1.00$\pm$0.01} \\
UniVL (ft)          &              0.60$\pm$0.19 &              0.85$\pm$0.12 &              0.58$\pm$0.19 &              0.27$\pm$0.22 &              0.90$\pm$0.09 \\
VideoClip (ft)      &              0.59$\pm$0.23 &              0.84$\pm$0.13 &              0.62$\pm$0.19 &              0.26$\pm$0.21 &              0.95$\pm$0.04 \\
\bottomrule
\end{tabular}}\end{table}

\subsection{Implementation Details}
To ensure fairness to the original models, we use the official model implementations that were available with pre-trained weights with the same experimental setup as described in these works. These protocols vary between models and datasets. HowTo100M-MIL \cite{miech2020end} take video as input and split the temporal boundary of the passed video into a clip of 4 with 32 frames for YouCook2 and 16 frames for MSRVTT. They take text as input and embed each word using Word2Vec. VideoClip \cite{videoclip} and COOT \cite{COOT} use pre-extracted features from the pre-trained S3D-G \cite{xie2018rethinking} model provided by \cite{miech2020end} while UniVL \cite{univl} uses pre-extracted features from the same model but before the final layer resulting in a smaller embedding size. VideoClip and UniVL take text as raw input while COOT \cite{COOT} uses pre-extracted text features from BERT \cite{BERT}. FIT \cite{fit} splits a clip into 4 segments and randomly selects 1 frame from each. These details are summarized in Table \ref{tab:models}.
We also analyze some models on whether they are fine-tuned, pre-trained or trained from scratch based on the availability of code. In the original implementations, VideoClip, Howto100-MIL and UniVL are pre-trained on HowTo100M \cite{howto100m}, COOT was trained from scratch on MSRVTT, and FIT is pre-trained on CC3M \cite{CC3M} and Web2Vid \cite{fit}.
Evaluating models using only pre-trained weights are considered \textit{zero-shot} (ZS). FIT, VideoClip and UniVL were additionally \textit{fine-tuned} (FT). Models that are trained on the evaluation datasets without pre-training are considered \textit{scratch}.

\section{Experiments}
\label{sec:experiments}

We perform our experiments with the studied models on YouCook2-P and MSRVTT-P benchmarks. A summarized overview of the robustness analysis of models against different perturbations on YouCook2-P is shown in Figure \ref{fig:teaser_yc2}. Table \ref{tab:overall_robustness} shows robustness scores aggregated across visual or real-world text perturbations for YouCook2-P and MSRVTT-P respectively. Table \ref{tab:visual_robustness_scores_msrvtt} show relative robustness scores aggregated across visual categories for MSRVTT-P. Table \ref{tab:text_perturbations} shows relative robustness scores across different text categories for both datasets. More detailed results, including a breakdown of each perturbation category, are provided in the Supplementary. Next, we provide more insights and analysis on different interesting observations in this study.
\begin{table}
    \centering
    \caption{Relative robustness scores $\gamma^r$ with standard deviations $\pm \sigma$ for each category of distribution shifts for text perturbations. 
    }
    \label{tab:text_perturbations}
    \resizebox{\textwidth}{!}{\begin{tabular}{{lllllllll}}
    \toprule
        \textbf{MSRVTT}$\gamma^r$ &                 AddText &                       Bias &                 ChangeChar &                   DropText &              Positional &                   SwapText &                  TextStyle \\
        \hline
FIT (scratch)       &              0.92$\pm$0.03 &              0.84$\pm$0.07 &              0.78$\pm$0.11 &              0.47$\pm$0.34 &              0.77$\pm$0.16 &              0.80$\pm$0.16 &  \underline{0.98$\pm$0.02} \\
VideoClip (scratch) &              0.90$\pm$0.06 &              0.88$\pm$0.05 &              0.78$\pm$0.09 &              0.46$\pm$0.32 &              0.73$\pm$0.13 &              0.81$\pm$0.18 &              0.96$\pm$0.02 \\
FIT (zs)            &     \textbf{1.00$\pm$0.00} &  \underline{0.96$\pm$0.04} &              0.79$\pm$0.14 &  \underline{0.53$\pm$0.36} &  \underline{0.84$\pm$0.13} &     \textbf{0.87$\pm$0.18} &     \textbf{1.01$\pm$0.02} \\
MIL NCE (zs)        &              0.78$\pm$0.00 &              0.90$\pm$0.03 &              0.77$\pm$0.10 &     \textbf{0.57$\pm$0.32} &              0.78$\pm$0.15 &              0.75$\pm$0.12 &              0.91$\pm$0.02 \\
UniVL (zs)          &              0.92$\pm$0.10 &     \textbf{0.97$\pm$0.04} &              0.71$\pm$0.11 &              0.33$\pm$0.27 &              0.64$\pm$0.15 &  \underline{0.84$\pm$0.15} &              0.90$\pm$0.07 \\
VideoClip (zs)      &              0.89$\pm$0.07 &              0.94$\pm$0.05 &              0.71$\pm$0.11 &              0.39$\pm$0.27 &              0.62$\pm$0.17 &              0.81$\pm$0.13 &              0.97$\pm$0.03 \\
FIT (ft)            &  \underline{0.94$\pm$0.04} &              0.88$\pm$0.05 &              0.79$\pm$0.11 &              0.49$\pm$0.34 &              0.80$\pm$0.14 &              0.82$\pm$0.17 &              0.97$\pm$0.02 \\
UniVL (ft)          &              0.92$\pm$0.05 &              0.88$\pm$0.05 &  \underline{0.80$\pm$0.09} &              0.49$\pm$0.31 &              0.78$\pm$0.13 &              0.81$\pm$0.14 &              0.96$\pm$0.01 \\
VideoClip (ft)      &  \underline{0.94$\pm$0.04} &              0.91$\pm$0.05 &     \textbf{0.81$\pm$0.09} &  \underline{0.53$\pm$0.32} &     \textbf{0.87$\pm$0.08} &              0.83$\pm$0.15 &              0.97$\pm$0.02 \\
\bottomrule
        \toprule
        \textbf{YouCook2}$\gamma^r$ &                AddText & Bias &                 ChangeChar &                   DropText &                 Positional &                   SwapText &               TextStyle \\

        \midrule
       COOT (scratch)      &              0.88$\pm$0.12 &  --- &              0.18$\pm$0.29 &              0.41$\pm$0.37 &              0.76$\pm$0.12 &              0.51$\pm$0.43 &              0.57$\pm$0.51 \\
VideoClip (scratch) &              0.85$\pm$0.12 &  --- &              0.62$\pm$0.13 &              0.37$\pm$0.33 &              0.69$\pm$0.11 &              0.72$\pm$0.18 &              0.92$\pm$0.05 \\
MIL NCE (zs)  &              0.92$\pm$0.03 &  --- &              0.74$\pm$0.14 &     \textbf{0.57$\pm$0.39} &     \textbf{0.83$\pm$0.15} &              0.75$\pm$0.18 &  \underline{0.98$\pm$0.01} \\
UniVL (zs)    &     \textbf{1.14$\pm$0.03} &  --- &              0.75$\pm$0.10 &              0.43$\pm$0.41 &              0.80$\pm$0.24 &              0.75$\pm$0.17 &              0.94$\pm$0.09 \\
VideoClip (zs)      &  \underline{0.95$\pm$0.04} &  --- &  \underline{0.77$\pm$0.10} &              0.47$\pm$0.33 &              0.70$\pm$0.13 &              0.77$\pm$0.14 &              0.94$\pm$0.04 \\
UniVL (ft)    &              0.91$\pm$0.08 &  --- &              0.74$\pm$0.09 &              0.45$\pm$0.33 &              0.76$\pm$0.07 &  \underline{0.78$\pm$0.15} &              0.95$\pm$0.02 \\
VideoClip (ft)      &  \underline{0.95$\pm$0.03} &  --- &     \textbf{0.84$\pm$0.10} &  \underline{0.50$\pm$0.35} &  \underline{0.82$\pm$0.09} &     \textbf{0.81$\pm$0.18} &     \textbf{0.99$\pm$0.07} \\
\bottomrule
        \end{tabular}}
    
\end{table}

\begin{table}
    \centering
    \caption{The aggregated performance measured by Relative Robustness $\gamma^r$ and Absolute robustness scores $\gamma^a$ across model and training procedure with standard deviations $\pm \sigma$. For text, we aggregated only natural distribution shifts, excluding Positional and DropText perturbations. 
    }
    \label{tab:overall_robustness}
     \resizebox{\textwidth}{!}{
\begin{tabular}{l|cc|cc|cc|cc}
\toprule 
\multirow{3}{*}{Method} & \multicolumn{4}{c|}{MSRVTT-P} & \multicolumn{4}{c}{YouCook2-P} \\

 &     \multicolumn{2}{c|}{Video}   &  \multicolumn{2}{c|}{Text} &                \multicolumn{2}{c|}{Video}   &  \multicolumn{2}{c}{Text}  \\
  &  $\gamma^a$     &     $\gamma^r$   &   $\gamma^a$  &   $\gamma^r$  &    $\gamma^a$   &    $\gamma^r$ &     $\gamma^a$   &     $\gamma^r$  \\
\hline
COOT (scratch)      &                        --- &                        --- &                     --- &                        --- &              0.79$\pm$0.16 &           0.52$\pm$0.36 &              0.75$\pm$0.19 &              0.44$\pm$0.44 \\
FIT (scratch)       &              0.93$\pm$0.08 &              0.84$\pm$0.18 &           0.94$\pm$0.05 &              0.87$\pm$0.11 &                        --- &                     --- &                        --- &                        --- \\
VideoClip (scratch) &              0.83$\pm$0.15 &              0.63$\pm$0.32 &           0.94$\pm$0.04 &              0.87$\pm$0.10 &  \underline{0.86$\pm$0.11} &           0.53$\pm$0.35 &  \underline{0.95$\pm$0.05} &              0.83$\pm$0.18 \\
FIT (zs)            &     \textbf{0.96$\pm$0.06} &     \textbf{0.91$\pm$0.15} &  \textbf{0.97$\pm$0.05} &     \textbf{0.92$\pm$0.13} &                        --- &                     --- &                        --- &                        --- \\
MIL NCE (zs)        &              0.89$\pm$0.08 &              0.60$\pm$0.29 &           0.96$\pm$0.02 &              0.85$\pm$0.09 &              0.84$\pm$0.13 &           0.53$\pm$0.37 &  \underline{0.95$\pm$0.05} &              0.86$\pm$0.15 \\
UniVL (zs)          &  \underline{0.94$\pm$0.05} &              0.67$\pm$0.30 &  \textbf{0.97$\pm$0.02} &              0.85$\pm$0.14 &     \textbf{0.91$\pm$0.07} &           0.50$\pm$0.36 &     \textbf{0.98$\pm$0.03} &  \underline{0.88$\pm$0.17} \\
VideoClip (zs)      &              0.92$\pm$0.07 &              0.66$\pm$0.32 &  \textbf{0.97$\pm$0.03} &              0.86$\pm$0.14 &              0.74$\pm$0.19 &           0.50$\pm$0.37 &              0.93$\pm$0.06 &              0.86$\pm$0.11 \\
FIT (ft)            &              0.92$\pm$0.09 &  \underline{0.86$\pm$0.15} &           0.93$\pm$0.06 &              0.88$\pm$0.10 &                        --- &                     --- &                        --- &                        --- \\
UniVL (ft)          &              0.82$\pm$0.15 &              0.65$\pm$0.29 &           0.94$\pm$0.05 &              0.88$\pm$0.09 &              0.80$\pm$0.16 &  \textbf{0.55$\pm$0.36} &              0.93$\pm$0.05 &              0.85$\pm$0.12 \\
VideoClip (ft)      &              0.82$\pm$0.16 &              0.66$\pm$0.30 &           0.94$\pm$0.05 &  \underline{0.89$\pm$0.09} &              0.72$\pm$0.23 &  \textbf{0.55$\pm$0.37} &  \underline{0.95$\pm$0.06} &     \textbf{0.91$\pm$0.10} \\

\bottomrule
\end{tabular}}
\end{table}

\begin{figure}[t!]
    \centering
    \includegraphics[width=.88\linewidth]{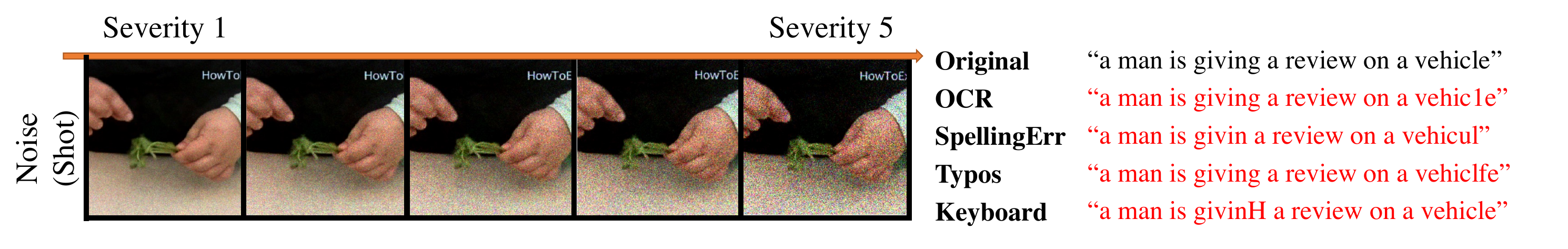}
    \caption{Examples of perturbations that humans are able to perceive but models struggle with.
    }
    \label{fig:human_perceive}
\end{figure}

\paragraph{Training Strategy}
Table \ref{tab:overall_robustness} split models by their training strategy. These results indicate that for MSRVTT-P, models that are zero-shot are typically higher in absolute and relative robustness. For long, complex activities in YouCook2-P, fine-tuned models are typically more relatively robust. Pre-training data choice may also play a factor. FIT pre-trains on both images and video as opposed to the majority of the other approaches that pre-train or use features pre-trained on the HowTo100M dataset \cite{howto100m}. While FIT performs well on MSRVTT, when zero-shot evaluating FIT on YouCook2 without perturbations, the results are an R@5 of $7.5\%$, indicating this may only be the case for short activity videos like in MSRVTT. 
\textit{In summary, pre-training models typically improves both performance and robustness against real-world and synthetic distribution shifts}.

\paragraph{Human Perceivable Perturbations}
\textit{Noise} and \textit{Blur} are pixel-based visual perturbations which humans can easily filter
(Fig. \ref{fig:human_perceive}). These perturbations are also ones models are least robust as shown for MSRVTT-P in Table \ref{tab:visual_robustness_scores_msrvtt}. While the models pre-trained on HowTo100M using a CNN backbone feature extractor perform poorly on spatial noise in MSRVTT-P, the ViT based approach FIT is more relatively robust to noise. On text perturbations for both datasets shown in Table \ref{tab:text_perturbations}, \textbf{between the semantic preserving} text perturbations, models are \textbf{least robust to ChangeChar}, indicating that text models are still unable to recognize small changes that humans will perceive in text (Fig. \ref{fig:human_perceive}).
\textit{ This indicates that visual-language models are \textbf{not typically robust to real-world distribution 
shifts that are human perceivable} such as character changes in word(s) and additive noise}. 

\paragraph{Architecture}
\begin{table}[t!]

\end{table}

There are typically two architecture types for video-language models, a two-branch or cross-attention encoder. Two-branch encoders keep the visual and text encoders separate with the only interaction being the propagated loss. Cross-attention utilizes a form of cross-attention between a visual and language encoder before calculating loss. Most models use a two-branch encoder approach as they find it performs better \cite{videoclip}. However, of the models we studied and evaluated on YouCook2-P, COOT \cite{COOT} and UniVL \cite{univl} use cross-attention while VideoClip \cite{videoclip} and MIL NCE \cite{howto100m} use two-branch. Looking at Table \ref{tab:overall_robustness}, UniVL typically has higher absolute robustness scores compared to the two-branch encoder based approaches. \textit{Based on the models studied, this indicates the cross-attention may improve performance on clips from long, complex activities that are less distinguishable, with little cost to robustness}.

The visual encoder architecture also varies for the different approaches. The majority of the models studied here use a 3D CNN for video feature extraction that is input into a transformer. However, the FIT \cite{fit} model uses a small set of frames input to a ViT. When looking at Table \ref{tab:visual_robustness_scores_msrvtt}, there is a noticeable relative robustness difference between FIT and the other approaches. This indicates that ViT transformers may be more relatively robust than CNN based approaches. However, the FIT zero-shot model does not perform well on YouCook2, with a baseline R@5 of 7.5\%, indicating that \textit{using a ViT may be highly robust against short activities but not necessarily long, complex activities}. 

\begin{figure}[t!]
    \centering
    \includegraphics[width=\linewidth]{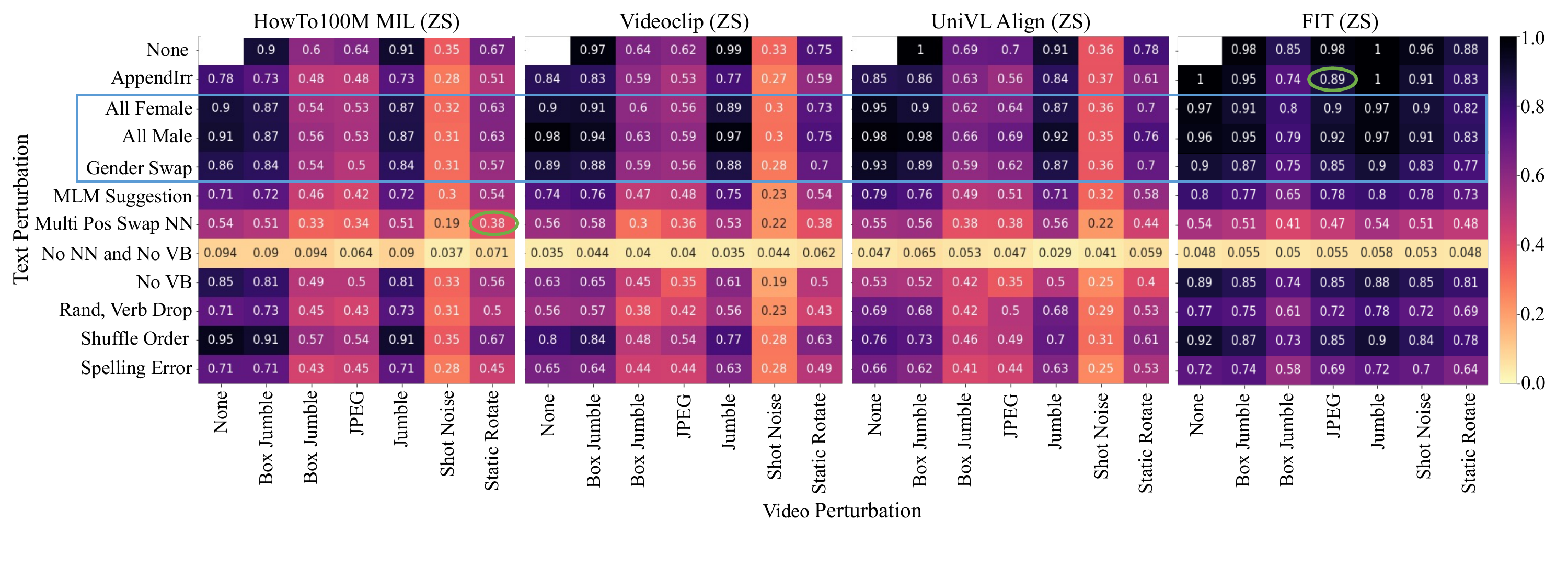}
    \caption{The drop from R@5 performance on clean to when perturbed by combinations of text and visual perturbations measured by Relative Robustness $\gamma^r$. The x-axis shows text perturbation and the y-axis shows visual perturbation. The first row/column show scores for the respective text/video perturbation when not combined. An example of compounding effect is circled in green. Gender perturbations are boxed in blue.} 
    \label{fig:multimodal_perturbations}
\end{figure}

\begin{figure*}[t!]
  \begin{center}
    \includegraphics[width=\textwidth]{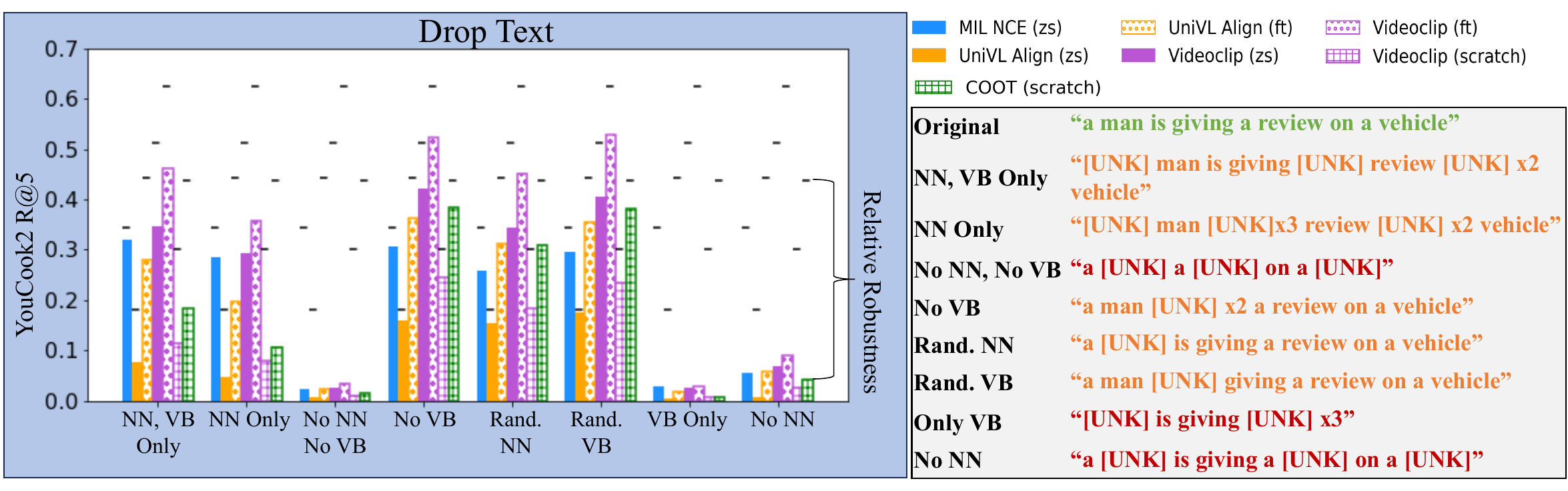}
  \end{center}
  \caption{Performance for \textit{DropText} perturbations on YouCook2-P. Dashes are R@5 on clean and bars are R@5 on perturbed. Examples of these perturbations are provided: red is where models struggle the most and orange indicates models are surprisingly robust.}
  \label{fig:drop_text}
\end{figure*}

Text encoders also vary across models. However, almost all approaches utilize a BERT \cite{BERT} transformer while only MIL NCE \cite{howto100m} use a Word2Vec \cite{word2vec}. When text is perturbed on \textit{DropText}, \textit{Positional} and \textit{ChangeChar}, Word2Vec is more robust than BERT on zero-shot evaluation (see Figure \ref{fig:drop_text}). 
\textit{Based on the models studied, these results may indicate that when using keywords as opposed to sentence descriptions, Word2Vec may be a more robust approach compared to BERT. } Additionally, as shown in Figure \ref{fig:teaser_yc2}, COOT's relative robustness is noticeably worse when text is perturbed. Because COOT uses pre-extracted text features without any training, this may indicate that using pre-extracted text features is not a robust method due to learning pulling the video feature space towards the existing text space rather than pulling both video and text closer together in a new, joint space.

\begin{figure}[t!]
    \centering
    \includegraphics[width=\linewidth]{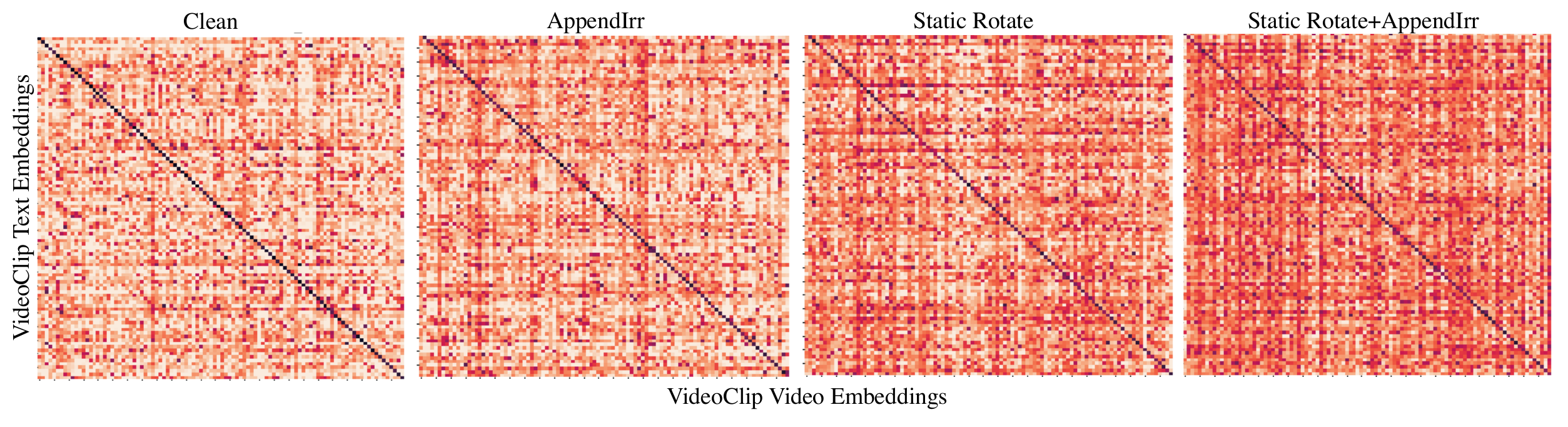}
    \caption{Similarity matrices where the x-axis are video representations and the y-axis are text representations sampled from VideoClip on the YouCook2 dataset. The darker the color, the more similar. When both video and text are perturbed, a compounding effect is shown by the increase in similarity for samples that do not match. } 
    \label{fig:compounding_effects}
\end{figure}

\paragraph{MultiModal Perturbations}
To understand the compounding effects of shifting distributions in both the visual and text domain, we select a subset from each perturbation with at least one from each higher-level category. For visual perturbations, we use a severity of 3. Figure \ref{fig:multimodal_perturbations} shows a summary of these results on the MSRVTT dataset. There are certain combinations of perturbations that are more challenging for models as compared to others. For example with FIT (ZS), the model has a relative robustness score $\gamma^r=1$ for \textit{AppendIrr} and $\gamma^r=0.98$ for \textit{JPEG} compression, but when combined, the score is $\gamma^r=0.89$ (see Figure \ref{fig:compounding_effects}). 

\begin{figure}[t!]
  \begin{center}
    \includegraphics[width=.85\textwidth]{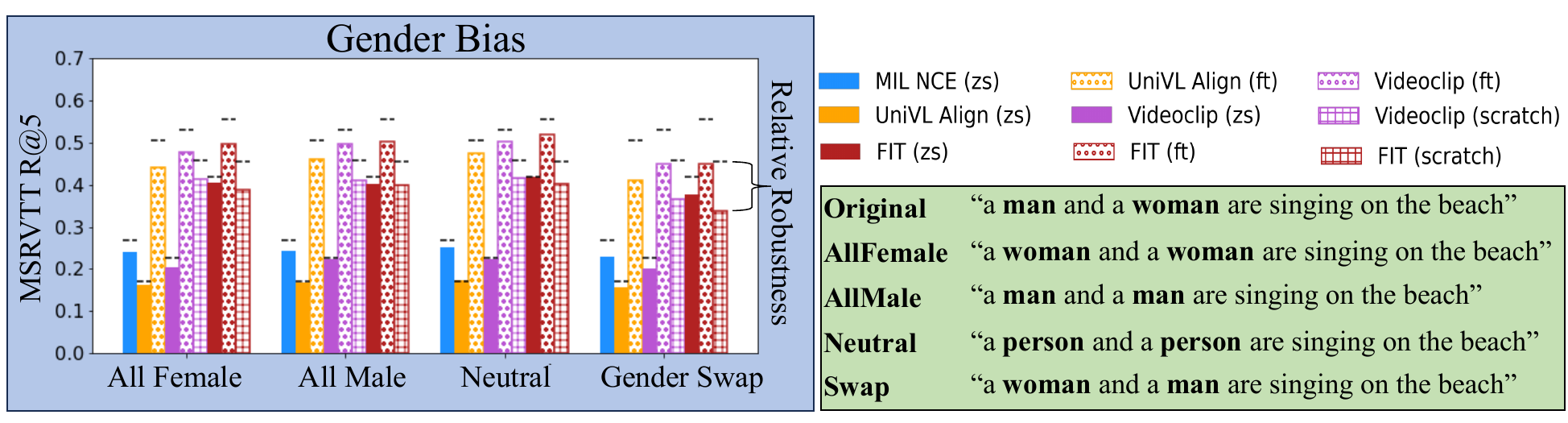}
  \end{center}
  \caption{Performance on \textit{Bias} perturbations on MSRVTT. Dashes are R@5 on clean and bars are R@5 on perturbed. Models are less robust when gender is swapped or male/female terms are changed. 
  } 
  \label{fig:bias_perts}
\end{figure}

Meanwhile, some perturbation combinations will be close to the lowest $\gamma^r$ between the two, e.g. no nouns and no verbs and shot noise. Even when a model is equally robust to perturbations in isolation (e.g. Jumble and GenderMale on HowTo100m MIL), there is a decrease in overall robustness when combined. 
In summary, \textit{when both text and video are perturbed, models are less robust than when the same perturbations are applied in isolation, with some combinations worse than others.}  

\paragraph{Bias}
\label{sec:bias_analysis}
To evaluate bias in models, we evaluate the robustness to gender-specific changes to text on the MSRVTT dataset. 
In the MSRVTT dataset, the most common part-of-speech (POS) tagged nouns were ``man'' and ``woman'' with ``man'' references 2x that of ``woman''. When the original text was perturbed, $33.8\%$ of male references were converted to female, $24.3\%$ of female references were converted to male, $53.2\%$ of phrases swapped gender and finally $52.8\%$ of gender references were made neutral. Figure \ref{fig:bias_perts} visualizes these results where the dotted, horizontal line is the original text-to-video retrieval score and the bar are the new scores with the perturbed version of text.\textit{ The results indicate that the selected models are \textit{less robust when the gender is all female} and when the gender is swapped from male-to-female and vice versa}. Additionally, when a caption is changed to gender neutral and evaluated zero-shot, models are very robust. However, when models are fine-tuned or trained from scratch, their performance has a noticeable drop. \textit{This further indicates models are learning gender biases to certain activities, likely as a result of biases in the target dataset they are trained on}.

\begin{figure*}[h]
  \begin{center}
    \includegraphics[width=.90\textwidth]{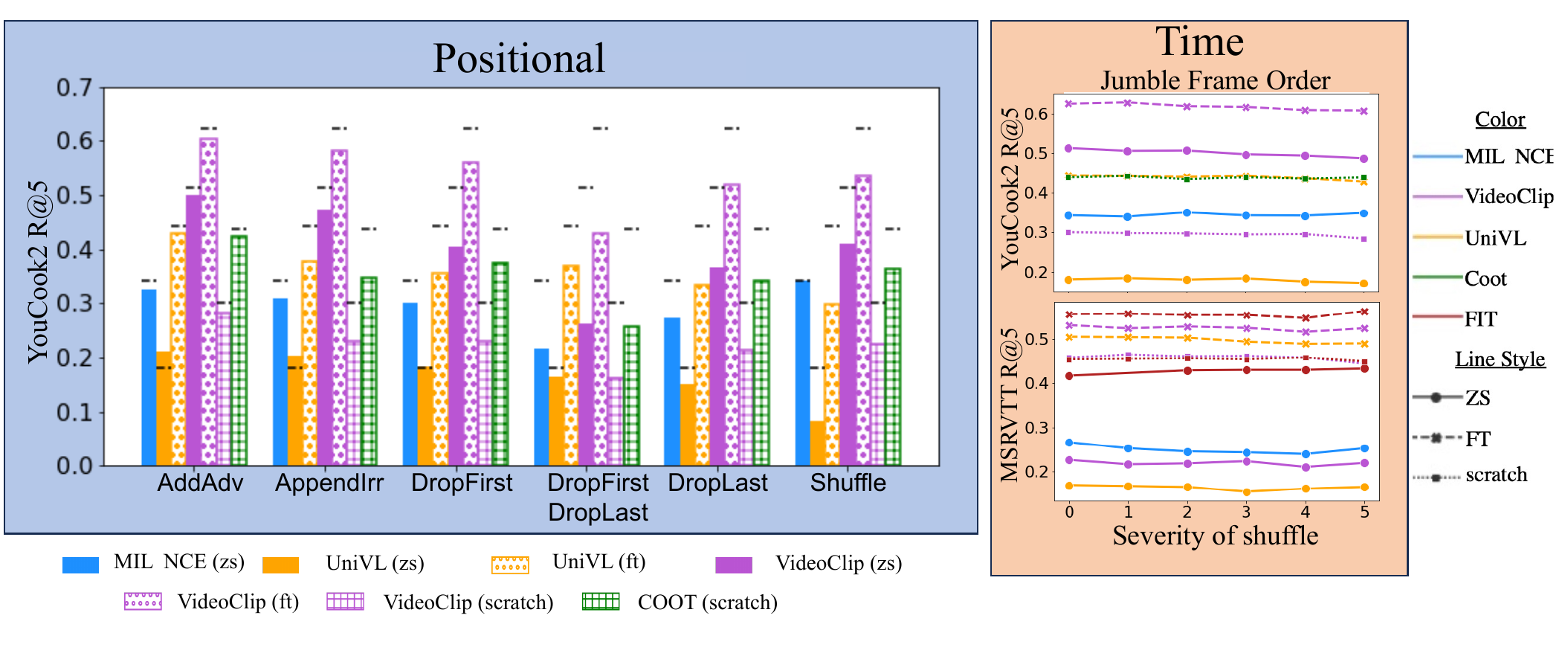}
  \end{center}
  \caption{Temporal perturbations for text (left) and video (right). On text, dashes are R@5 on clean and bars are R@5 on perturbed. Models appear to be typically robust, especially MIL NCE for \textit{ShuffleOrder}. For video, the x-axis is the severity where a severity of 0 is performance on clean and y-axis is R@5. Models show little change in performance, indicating that temporal order is not utilized in these approaches.}
  \label{fig:temporal_perts}
\end{figure*}

\paragraph{Temporal}
\label{sec:temporal_analysis}
Temporal perturbations are used to evaluate whether models use temporal information or not. Figure \ref{fig:temporal_perts} visualizes the results of these experiments. 
Models show strong robustness to the video-specific temporal perturbations \textit{Jumble}, \textit{Sampling}, and \textit{Freeze} (a breakdown of robustness scores is provided in the Supplementary). This indicates that fine-grained temporal elements are not necessarily important to these video-language models. This also indicates that the activities do not necessarily change when in reverse. 

On YouCook2-P, which consists of untrimmed, minutes-long videos, none of the models are robust to \textit{BoxJumble}. This indicates that the models require alignment between the visual ques and the respective text, but temporal order within the aligned segment is not utilized. While this shows poor robustness for this perturbation, it shows good model behavior. \textit{These results indicate that both visual and textual cues are used during learning however the \textbf{ models are attending more to objects and scene rather than motion and activity}.} This is similar to how humans may describe different videos where nouns and descriptors are more differentiating as opposed to activities which could describe a group of videos. More details on this conclusion in with regards to text is in the Supplementary E.5.

\section{Conclusion}
In this work we propose a robustness benchmark for video-language models and provide initial insights of several multimodal approaches.
In order to perform this study, we create two benchmark datasets, MSRVTT-P and YouCook2-P. Our empirical study provides several interesting insights into the behavior of some of the existing models on the proposed benchmarks. Some key observations are 1) models are generally more susceptible when only video is perturbed as opposed to when only text is perturbed, 2) models that are pre-trained are typically more robust with improved performance on zero-shot evaluation 3) models attend more to scene and objects rather than to motion and action. For example, original videos were often still matched when the gender is swapped in the query text, words are shuffled, and adverbs and adjectives swapped. 4) Models showed gender biases when trained on a target dataset as opposed to a larger, pre-training dataset. 
The findings and the benchmark in this work can potentially open up interesting future research on robustness of video-language learning. 

\bibliographystyle{plain}
\bibliography{mybib}

\appendix

\section{Implementation Details}
\subsection{Visual Perturbations}
Below are more details on the perturbations applied to videos. Examples of these perturbations can also be found at \url{https://bit.ly/3CNOly4}.

\paragraph{Noise}
These perturbations apply transformations at the pixel level of each frame in a video. The different noises are \textit{Impulse}, \textit{Gaussian}, \textit{Shot}, and \textit{Speckle}. \textbf{Impulse} noise simulates corruptions cause by bit errors by applying a combination of salt and pepper noise with amounts ranging from .03, .06, .09, 0.17, 0.27. \textbf{Gaussian} noise simulates low-lighting conditions by first normalizing the pixel values then adds a random normal noise scaled at values .08, .12, 0.18, 0.26, 0.38 based on severity. \textit{Shot} noise simulates electronic noise caused by the discrete nature of light by applying a combination of salt and pepper noise with amounts ranging from .03, .06, .09, 0.17, 0.27. \textit{Speckle} noise simulates additive noise and is similar to Gaussian but where the random value is then multiplied by the normalized pixel value.

\paragraph{Blur}
Blur perturbations apply transformations that simulate camera motion and focus. \textit{Motion} blur increases the radius and sigma of the kernel which is used to create the motion blurring effect ranging from (10, 3), (15, 5), (15, 8), (15, 12),  and (20, 15) based on severity. \textit{Zoom} blur blurs towards the center of the frame while increasing the zoom factor based on severity. \textit{Defocus} blur imitates a defocused lens over the entire frame. We increase the radius of the disk which is convolved over the image to create defocus blurring effect ranging from (3, 0.1), (4, 0.5), (6, 0.5), (8, 0.5), (10, 0.5) based on severity.

\begin{figure}[t!]
    \centering
    \includegraphics[width=.5\linewidth]{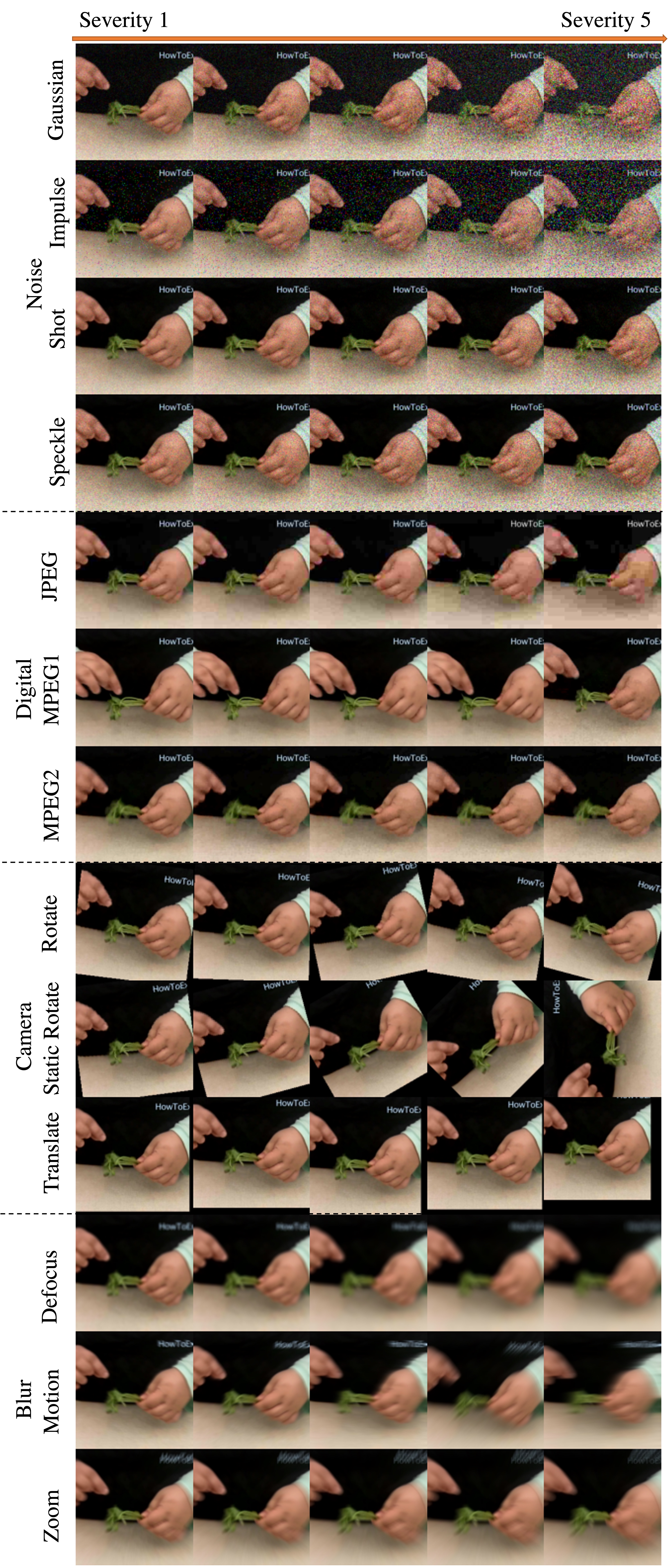}
    \caption{Visualizations of each perturbation for a single frame in a video from the YouCook2 dataset. Severity increases from left to right for each perturbation. }
    \label{fig:image_grid}
\end{figure}

\paragraph{Digital} 
\textit{JPEG} compression converts each frame to a JPEG with quality ranging from 25, 18, 15, 10, 7 based on severity. \textit{MPEG1} compresses the original video to using the ffmpeg \cite{ffmpeg} format mpeg2video with levels ranging from 20, 40, 60, 80, 100. \textit{MPEG2} compresses the original video to using the ffmpeg format mpeg4 with levels ranging from 15, 30, 45, 60, 75. This compression tends to actually affect the playing of the video, where frames are missing and/or skipped. These slight frame changes allows these perturbations to be considered temporal as well. This can be seen in an example in Figure \ref{fig:image_grid} under Digital where the frame does not perfectly align with the frames for the other perturbations because it is slightly off temporally. We can consider these perturbations as spatio-temporal as they alter both spacial features and temporal features.

\paragraph{Temporal}
\textit{Jumbling} splits a video into segments of lengths 32, 16, 8, 4, and 2 where the higher is less severe and lower is more severe. The frames within each segment are then randomly shuffled. \textit{Box Jumbling} splits a video into segments of lengths 4, 9, 16, 25, 36 where the higher is more severe and lower is less severe. The segments are then randomly shuffled. \textit{Sampling} transforms a video from the original method’s frames per second (FPS) to keep consistency with the original approach. However, it slows the playback speed by sampling frames uniformly at a varying level of rates 2, 4, 8, 16, and 32, where the higher is more severe. \textit{Reverse Sampling} is the same as sampling but also reverses the video after sampling. \textit{Freeze} This perturbation will choose a percentage of frames to select, ranging from 40\%, 20\%, 10\%, 5\% and 2.5\%. The more frames selected, the less severe the perturbation. These selected frames are then repeated until they reach the next sequential frame, simulating a frozen live stream video. 

\paragraph{Camera}
These perturbations simulate irregularities with camera motion and include \textit{Static rotation}, \textit{Rotation} and \textit{Translation}. Static rotate rotates every frame the same degree, Rotation rotates each frame by a random degree, and Translation randomly chooses a new center in the frame to crop to for each frame as if the camera is randomly shaking.

\begin{figure}[t!]
    \centering
    \includegraphics[width=.85\textwidth]{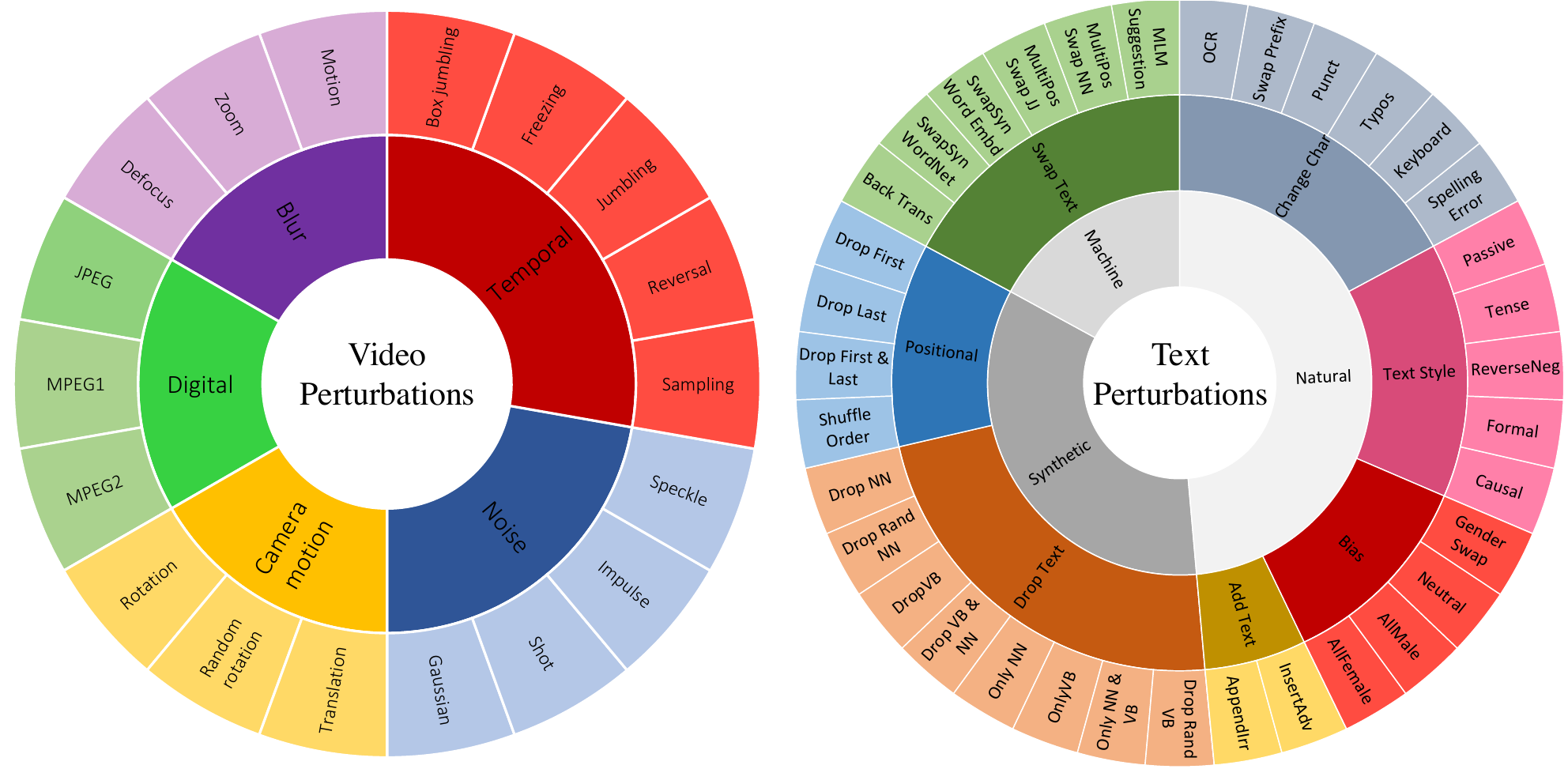} 
    \caption{The summary of how this paper organizes both visual and text perturbations used to evaluate on the text-to-retrieval task for multimodal models. On the visual side, each perturbation also ranges from a severity of 1 to 5.}
    \label{fig:video_perturbations}
\end{figure}

\begin{figure}[t!]
    \centering
    \includegraphics[width=.99\linewidth]{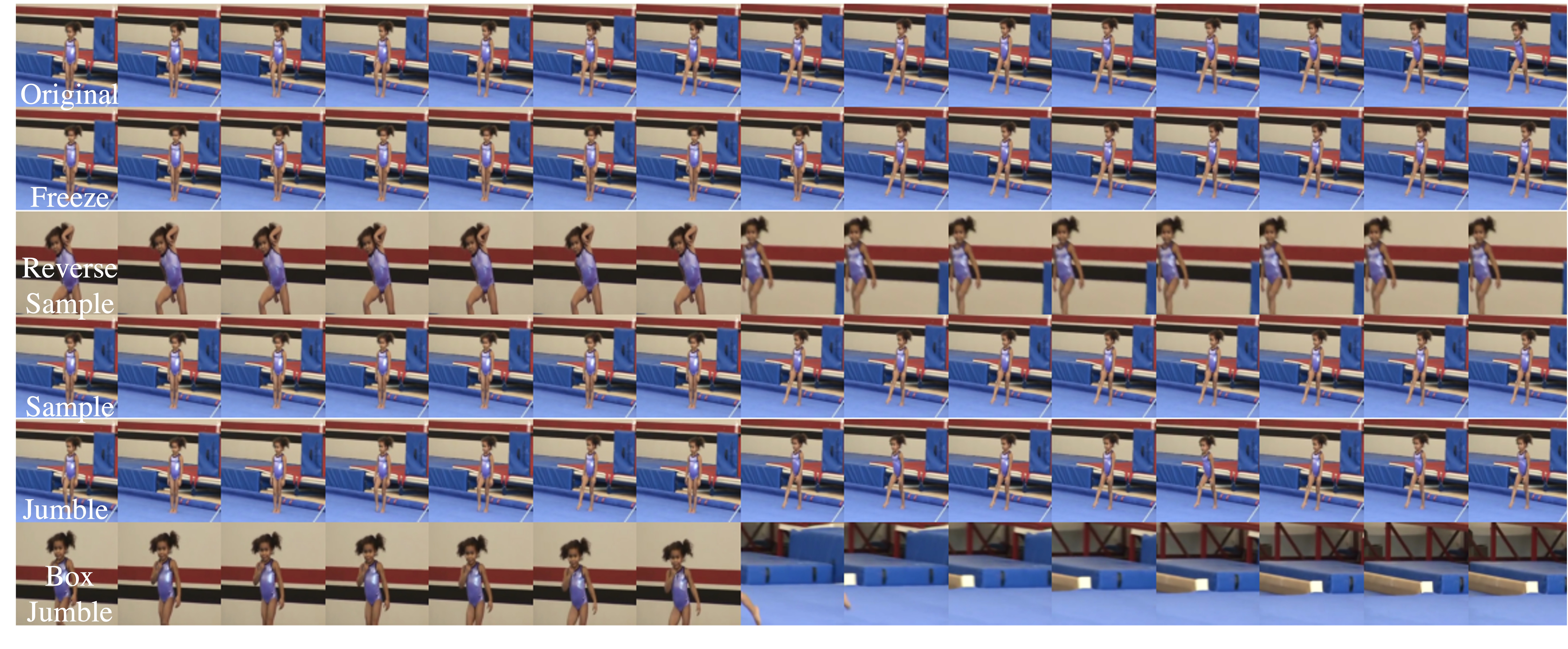}
    \caption{A visualization of the temporal perturbations for a video showing ``a little girl does gymnastics''. }
    \label{fig:my_label}
\end{figure}

\begin{table}[t!]
    \centering
    \caption{Examples of all text perturbations in each category fot the text ``a little girl does gymnastics'' from the MSRVTT dataset.}
    \label{tab:text_examples}
\begin{tabular}{lll}
\toprule
      Type &  Perturbation &                                       Perturbed \\
\midrule
   \multirow{2}{*}{AddText} & AddAdv  &       a little girl specifically does gymnastics \\
          & AppendIrr  &  On this occasion, a little girl does gymnastics \\
          \hline
\multirow{4}{*}{Bias} & AllFemale  &                    a little girl does gymnastics \\
          & AllMale  &                     a little boy does gymnastics \\
          & GenderNeutral  &                   a little child does gymnastics \\
          & GenderSwap  &                     a little boy does gymnastics \\
          \hline
\multirow{6}{*}{ChangeChar} & Keyboard  &                    a little girl dofs gymnastics \\
          & OCR  &                    a little girl does gymnastic8 \\
          & PrefixSwap  &                    a little girl does gymnastics \\
          & Punct  &               " a little girl does gymnastics, " \\
          & SpellErr  &                    a lettil girl does gymnastics \\
          & Typos  &                     a little girl des gymnastics \\
          \hline
\multirow{8}{*}{DropText} & NN\&VBOnly  &                 [UNK] [UNK] girl does gymnastics \\
          & NNOnly  &                [UNK] [UNK] girl [UNK] gymnastics \\
          & NoNN\&VB  &                       a little [UNK] [UNK] [UNK] \\
          & NoVB  &                   a little girl [UNK] gymnastics \\
          & RandNN  &                         a little girl does [UNK] \\
          & RandVB  &                   a little girl [UNK] gymnastics \\
          & VBOnly  &                     [UNK] [UNK] [UNK] does [UNK] \\
          & NoNN  &                        a little [UNK] does [UNK] \\
          \hline
\multirow{4}{*}{Positional} & DropFirst  &                [UNK] little girl does gymnastics \\
          & DropFirstLast  &                     [UNK] little girl does [UNK] \\
          & DropLast  &                         a little girl does [UNK] \\
          & ShuffleOrder  &                    a girl gymnastics does little \\
          \hline
\multirow{6}{*}{SwapText} & BackTrans  &                         a little girl gymnastics \\
          & JJSwap  &                   a anodyne girl does gymnastics \\
          & MLM  &                   a teenage girl does gymnastics \\
          & NNSwap  &                  a little output does gymnastics \\
          & SynWordEmbd  &                      a good girl does gymnastics \\
          & SynWordNet  &                  a little girl manage gymnastics \\
          \hline
\multirow{5}{*}{TextStyle} & Casual  &               A little girl that does gymnastics \\
          & Formal  &                   A young woman does gymnastics. \\
          & Neg  &                a little girl does not gymnastics \\
          & Passive  &              gymnastics is done by a little girl \\
          & Tense  &                     a little girl did gymnastics \\
\bottomrule
\end{tabular}
\end{table}

\subsection{Text Perturbations}
In Table \ref{tab:text_examples} there are examples for each perturbation when the input text is ``a little girl does gynmanstics'' from the MSRVTT dataset. This section discusses the implementation of each in more detail. 

\paragraph{ChangeChar}
Perturbations, natural and machine-based, that alter a character in one or several of the words in the text. For natural-based perturbations, this includes \textit{SpellingError}, \textit{Keyboard}, and \textit{Typos} \cite{textflint}. These are based on common spelling errors, keyboard mistypes, and general typos. Typos for example randomly inserts, deletes, swaps or replaces a single letter within one word while keyboard alters text by common keyboard mistakes such as ``word $\rightarrow$ to work''. Machine-based perturbations include \textit{OCR}, \textit{SwapPrefix} and \textit{Punctuation} \cite{textflint}. SwapPrefix for example will swap the prefix of one word while keeping its part-of-speech tag. Punctuation appends and/or prepends random punctuation to the sentence and OCR uses random values to stimulate an OCR, or optical character recognition, error. 

\paragraph{SwapText}
It is machine-based perturbations that swap word(s) from the original text. Several perturbations that make word swaps based on text models include \textit{BackTrans} which replaces text with phrases by using back-translation \cite{textflint}. \textit{SwapSynWordNet} and \textit{SwapSynWordEmbedding} both swap a words with their synonyms as determined by either WordNet \cite{wordnet} or by Glove \cite{glove}. \textit{MLM suggestion} swaps one syntactic category element of the sentence with what would be predicted by masked language models (MLM) \cite{textflint}. \textit{MultiPosSwapJJ} and \textit{MultiPosSwapNN} replaces adjectives and nouns respectively with words holding multiple parts-of-speech (POS).

\paragraph{AddText}
Perturbations \cite{textflint} that are natural-based and add text to the original. \textit{AppendIrr} appends irrelevant phrases to the original text while \textit{InsertAdv} adds an adverb before each verb. 

\paragraph{TextStyle}
Perturbations that are natural-based and change the style of the text. These include \textit{Tense}, \textit{Passive}, \textit{Casual}, \textit{Formal}, and \textit{ReverseNeg} \cite{styleformer}. The perturbations Passive, Casual, and Formal change the text style to those specific styles. Tense changes the tense of the text and ReverseNeg reverse negates the original text. 

\paragraph{Bias}
Perturbations that are natural-based that change the gender of a given phrase. These vary in \textit{AllFemale}, \textit{AllMale}, \textit{GenderSwap}, and \textit{Neutral}. The netural perturbation removes female and male references and replaces them with neutral ones. For example, a reference to ''a man'' and ''a woman'' would be replaced with ''a person''. GenderSwap swaps male references with female and vice versa using \cite{genderbender}.

\paragraph{DropText}
Perturbations are synthetic and drop words based on their part-of-speech (POS) tag. \textit{DropNN}, \textit{DropVB}, and \textit{DropVBNN} are different variations of dropping words based on whether the POS tags are Noun and/or Verb. \textit{OnlyNN}, \textit{OnlyVB}, and \textit{OnlyNNVB} drops all words but those with POS NN and/or VB. \textit{RandNN} and \textit{RandVB} drop one random noun/verb. This is done using the NLTK \cite{nltk} package to first extract POS tags for each word. Using these POS tags, based on the respective perturbation, words are ``dropped'' by replacing them with ``[UNK]'' in order to maintain the original phrase length. 

\paragraph{Positional}
These include \textit{DropFirst}, \textit{DropLast}, \textit{DropFirstandLast}, and \textit{ShuffleOrder}. Drop-related perturbations will replace a word at that position with an ``[UNK]'' tag. The ShuffleOrder perturbation shuffles the words in a phrase randomly. More details on the generated text perturbation are provided in the Appendix.

\subsection{Analysis of Perturbed Text}
To understand the severity of each perturbation we evaluate the perturbed text that is generated using multiple metrics including \textit{perplexity}, \textit{BLEU}, \textit{METEOR} and \textit{ROUGE}.

\begin{figure}[t!]
    \centering
    \includegraphics[width=.99\linewidth]{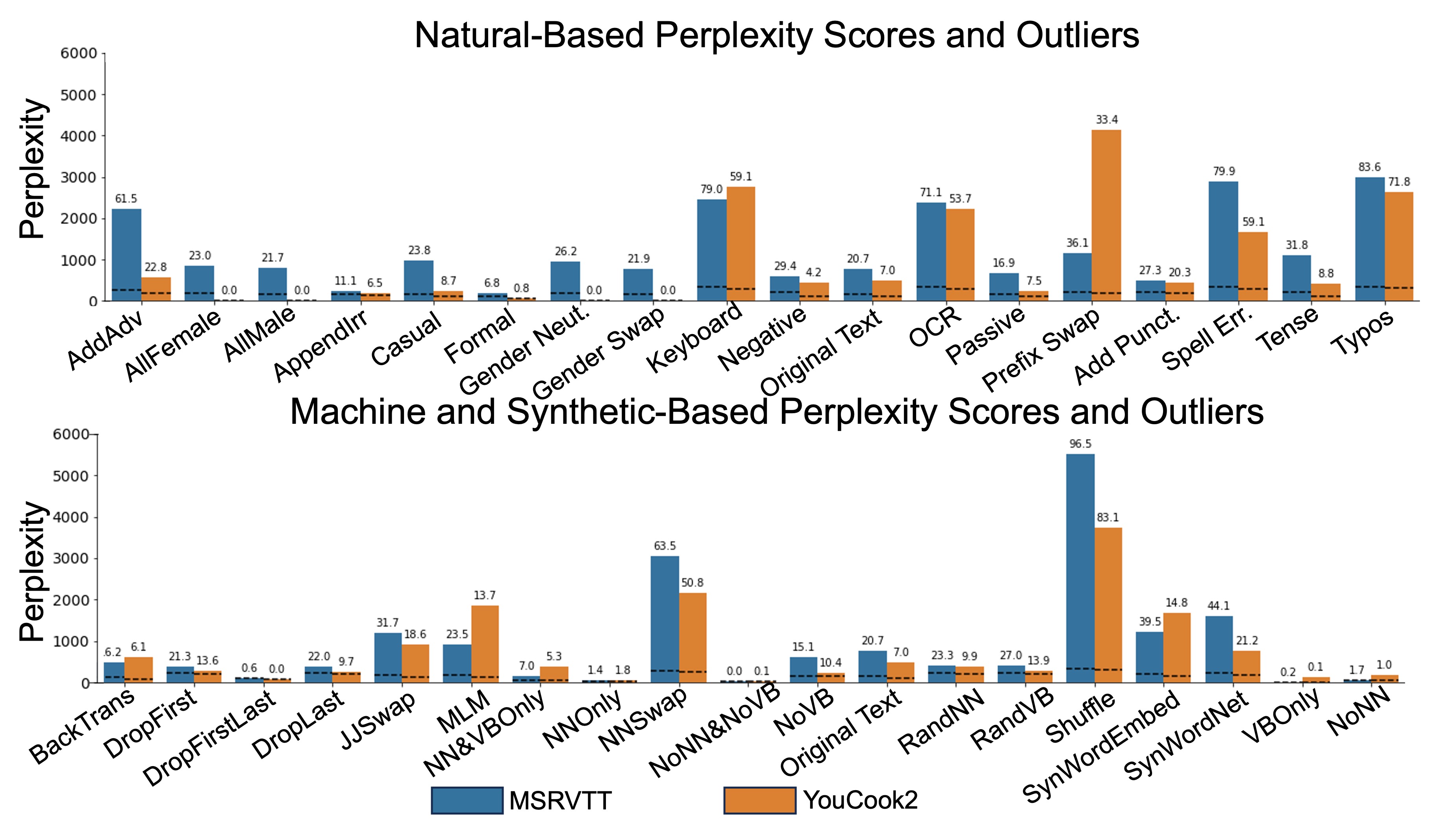} 
    \caption{The perplexity scores for the different text perturbations using GPT-3. The bars represent the average perplexity for the entire corpus, the dashed lines represent the perplexity when removing outliers based on a threshold of a 500, and the numbers atop the bars are the percent of outliers that are removed when using the threshold.}
    \label{fig:perplexity}
\end{figure}

\paragraph{Perplexity of Perturbations}
We first look at the perplexity measurement which measures the probability of a sentence using the large text model GPT-3 \cite{gpt3}. For each word in the sentence, the probability of the next word being present is measured and if the probability of the next word is low, then the perplexity for that sentence will be higher. Perplexity is not necessarily a good measurement for quality, but it is useful for measuring how statistical models may struggle with text. We first observe further support on how different the two datasets are on the text side where the perplexity of MSRVTT is 762.97 and for YouCook2 480.84. 

The results of this analysis is shown in Figure \ref{fig:perplexity} where machine-based perturbations are the bottom row and natural-based perturbations are the top. Between natural and machine-based there are no obvious differences in perplexity overall, both appear to have challenging distribution shifts according to the perplexity of GPT-3. Changing characters in words appear to result in higher perplexity consistently across the different variations of \textit{ChangeChar} perturbations. PrefixSwap and NNSwap are additionally high in perplexity on the machine-based side. These results would indicate the statistical models should struggle most with character changes and word swaps or drops based on POS-tags. The most perplex version of text is when words are shuffled in \textit{ShuffleOrder}, as the words positions to each other are no longer meaningful. In summary, machine-learning based approaches \textit{are likely to struggle most with character swapping perturbations and shuffling of words}.

\paragraph{Comparison Metrics to Original Text}
We also compare the perturbed text to the original text using the traditional metrics, BLEU \cite{papineni2002bleu}, METEOR and Rouge. The results for these are averaged across the different perturbations for each type for both the MSRVTT and YouCook2 datasets in Table \ref{tab:msrvtt_ppl}. Perturbations that \textit{DropText} are most dissimilar to the original text for both datasets. Depending on the dataset, \textit{AddText}, \textit{TextStyle} and \textit{ChangeChar} are similarly dissimilar to the original text, meaning models should be robust but show some level of performance reduction. The most similar is \textit{Bias}, meaning models should be highly robust to Bias.

In summary, these scores indicate that we have a varying level of difficulty with our text perturbations across categories, allowing for variable securities of distribution shift. The \textit{most challenging is DropText} and \textit{the least challenging is Bias}.

\begin{table}[t!]
    \centering
    \caption{Distribution Shift evaluation on MSRVTT and YouCook2 captions respectively. }
    \label{tab:msrvtt_ppl}
    \resizebox{.85\textwidth}{!}{\begin{tabular}{lrrrrrrrr}
        \toprule
        MSRVTT &  AddText &    Bias & ChangeChar & DropText &    Positional & SwapText & TextStyle \\
        \midrule
        BLEU4      &     0.57 &    0.88 &       0.60 &     0.29 &           0.64 &     0.68 &      0.56 \\
        Meteor     &     0.76 &    0.94 &       0.79 &     0.53 &          0.78 &     0.87 &      0.80 \\
        ROUGE-l F1  &     0.16 &    0.22 &       0.22 &     0.17 &        0.21 &     0.21 &      0.21 \\
        \bottomrule 
        \toprule
        YouCook2 &  AddText & Bias & ChangeChar & DropText & Positional & SwapText & TextStyle \\
        \midrule
        BLEU4      &     0.64 &  ----- &       0.58 &     0.34  &     0.62 &     0.66 &      0.65 \\
        Meteor     &     0.76 &  ----- &       0.78 &     0.52 &        0.76 &     0.85 &      0.81 \\
        ROUGE-l F1  &     0.17 &  ----- &       0.23 &     0.17 &       0.22 &     0.22 &      0.22 \\
        \bottomrule
        \end{tabular}}

\end{table}

\subsection{Model Implementations}
To process data, train and evaluate models, we used our internal cluster with single-GPU use per run. All models except the MIL NCE \cite{howto100m} and FIT \cite{fit} use features extracted from the visual encoder of the MIL NCE. For MIL NCE, the clips were split into 4 clips of 32 frames each. For VideoClip \cite{videoclip} and COOT \cite{COOT} input videos of size $224\times224$ at 30 fps were fed to the pre-trained S3D model where we extracted features at the final layer with an embedding size of 512. The same procedure was used for UniVL \cite{univl} with the difference being the features are extracted at the earlier layer ``mixed5c'' with an embedding size of 1024. For perturbations, we first perturb the video before extracting S3D features, therefore we have collect perturbed S3D features for embedding sizes 512 and 1024 for each perturbation and severity. We used the original code for these models to extract features to ensure that the procedure is the same as the original authors'. These original feature extraction scripts are located in the Github repositories \cite{feature_extract} and \cite{feature_extract_fairseq}. FIT \cite{fit} splits each clip into 4 segments and randomly selects a single frame from each segment.

\section{Limitations}
This study has several limitations which include 1) the use of simulated noise due to the challenge of obtaining real-world data and 2) the analysis is on a limited number of models due to availability of usable code and model weights. The analysis provided is limited to the models we have used in this study. We tried our best to benchmark all publicly available models that provided weights and used both text and video. Further challenges from each approach included 1) having strict requirements on large data pre-processing, 2) requiring heavy GPU usage for both testing and even more so for training, and 3) having their own specific testing bed. These factors limited both the selection and time it took to implement models. In future, hopefully a larger number of models will be available.


\section{Licensing}
All the models which we have used in this study are available in public domain. The model code for HowTo100m MIL \cite{howto100m} and COOT \cite{COOT} have the Apache 2.0 license and the model code for VideoClip \cite{videoclip} and UniVL \cite{univl} has the MIT license and are publicly available. We will provide YouCook2-P and MSRVTT-P publicly for future research. These datasets are based on existing YouCook2 \cite{youcook2} and MSRVTT \cite{msrvtt} datasets and we are not using any new video sources. Both these datasets are available in public domain for research purposes and therefore similar licensing is applicable to the newly curated datasets.

\section{Impact}
To our understanding, there are no negative societal impacts of our work. The goal of this work was to evaluate the robustness of models that may later be used in real-world settings. We aimed to improve the societal impacts by evaluating these models on real-world distribution shifts including potential bias in text.

\section{Additional Results}
\begin{figure}[t!] 
    \centering
    \includegraphics[width=.80\linewidth]{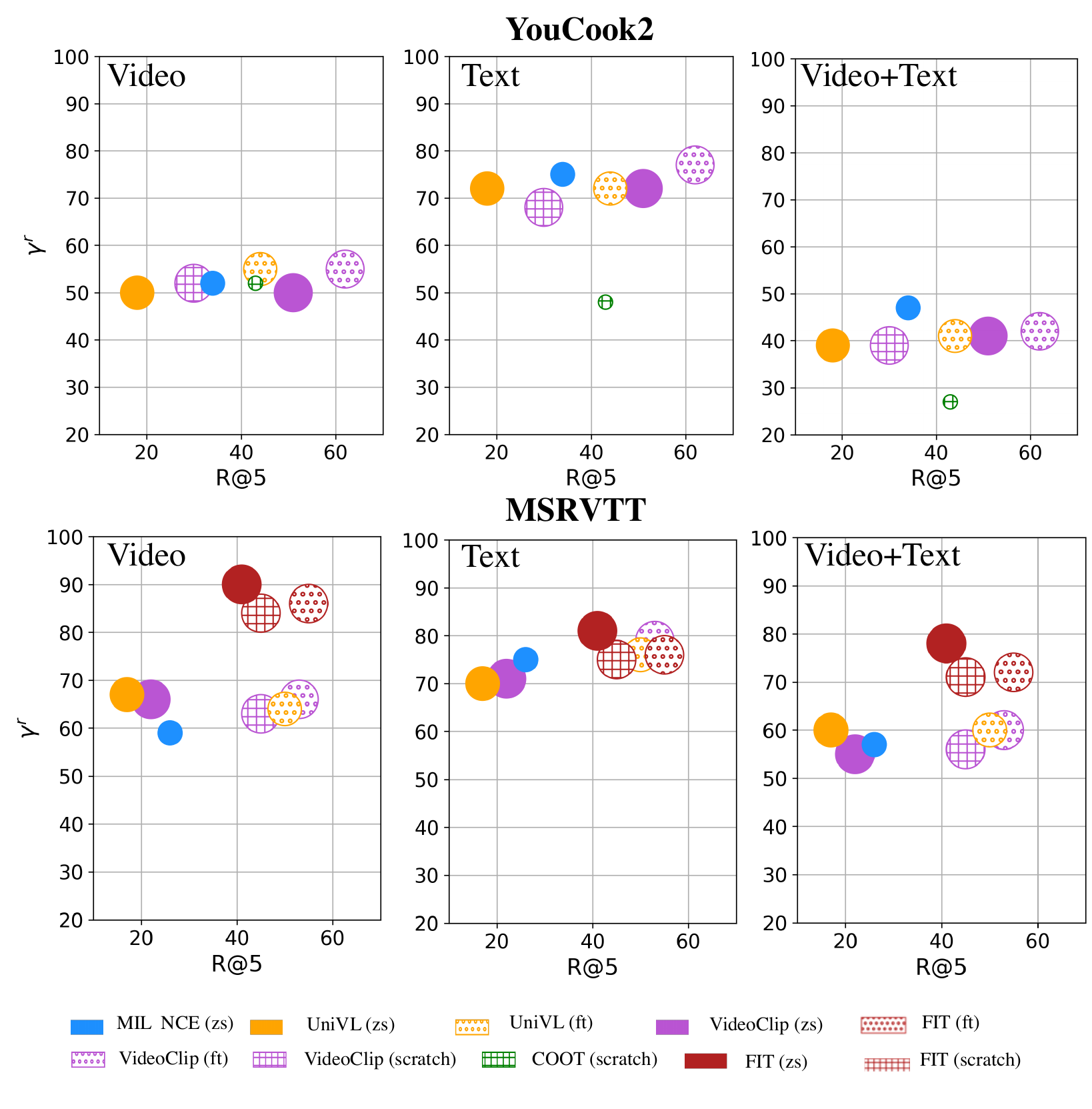}
    \caption{On the x-axis, R@5 for text-to-video retrieval and y-axis is the drop in performance when data is perturbed measured by relative robustness $\gamma^r$ for both datasets. These results are aggregated across all categories for the modality and all combinations we generated for the combination of video and text.}
    \label{fig:teaser_all}
\end{figure}

Figure \ref{fig:teaser_all} shows the relative robustness for the text-to-video retrieval task at R@5 aggregated across all categories for video, text and when both are perturbed against the performance. The results vary based on the dataset due to MSRVTT focusing on short videos of activity and YouCook2 breaking up a long-complex activity into shorter clips. The differences in model robustness and performance for the different datasets \textit{indicates a difference on how models handle clips from long, complex activities} compared to videos that are short and of a simple activity. On the longer activity dataset YouCook2, pre-training is noticeably more important for both robustness and performance. 

\subsection{Analysing Feature Space}
\begin{figure}[t!] 
    \centering
    \includegraphics[width=.99\linewidth]{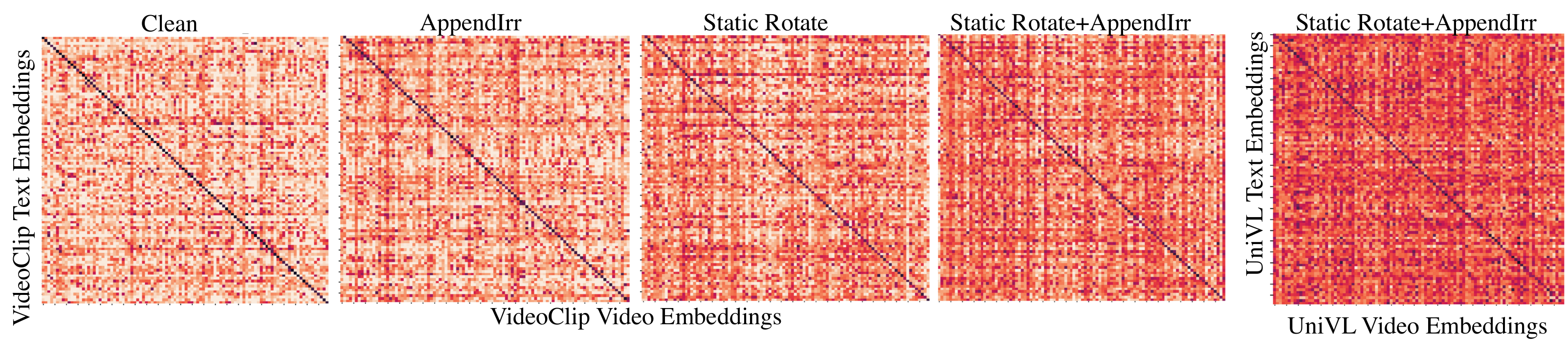}
    \caption{Similarity matrices where the x-axis are video representations and the y-axis are text representations sampled on the YouCook2 dataset. The darker the color, the more similar. When both video and text are perturbed, a compounding effect is shown by the increase in similarity for samples that do not match. Additionally, VideoClip shows less similarity between incorrect pairs when both are perturbed as compared to UniVL which utilizes cross-attention.}
    \label{fig:videoclip_compounding}
\end{figure}

To visualize the changes to the embedding space when text and video are perturbed, we selected videos that were accurately matched to their respective text in the baseline and observed their similarity change when perturbations were made. Figure \ref{fig:videoclip_compounding} visualizes the baseline, when text is perturbed with appending irrelevant phrases, when video is perturbed by a consistent rotation and when both these perturbations are applied. As these perturbations are added, the similarity between video and other text begins to increase. When both video and text are perturbed, this effect is most visible. Additionally, the UniVL model which uses cross-attention shows even more similarity between video and text when both are perturbed. This does not necessarily mean that UniVL is less robust in this case, but it can indicate with cross-attention, video and text are generally more similar and the difference between ranking one video to the other is much smaller than in other models. 

\begin{figure}[t!]
    \centering
    \includegraphics[width=.95\linewidth]{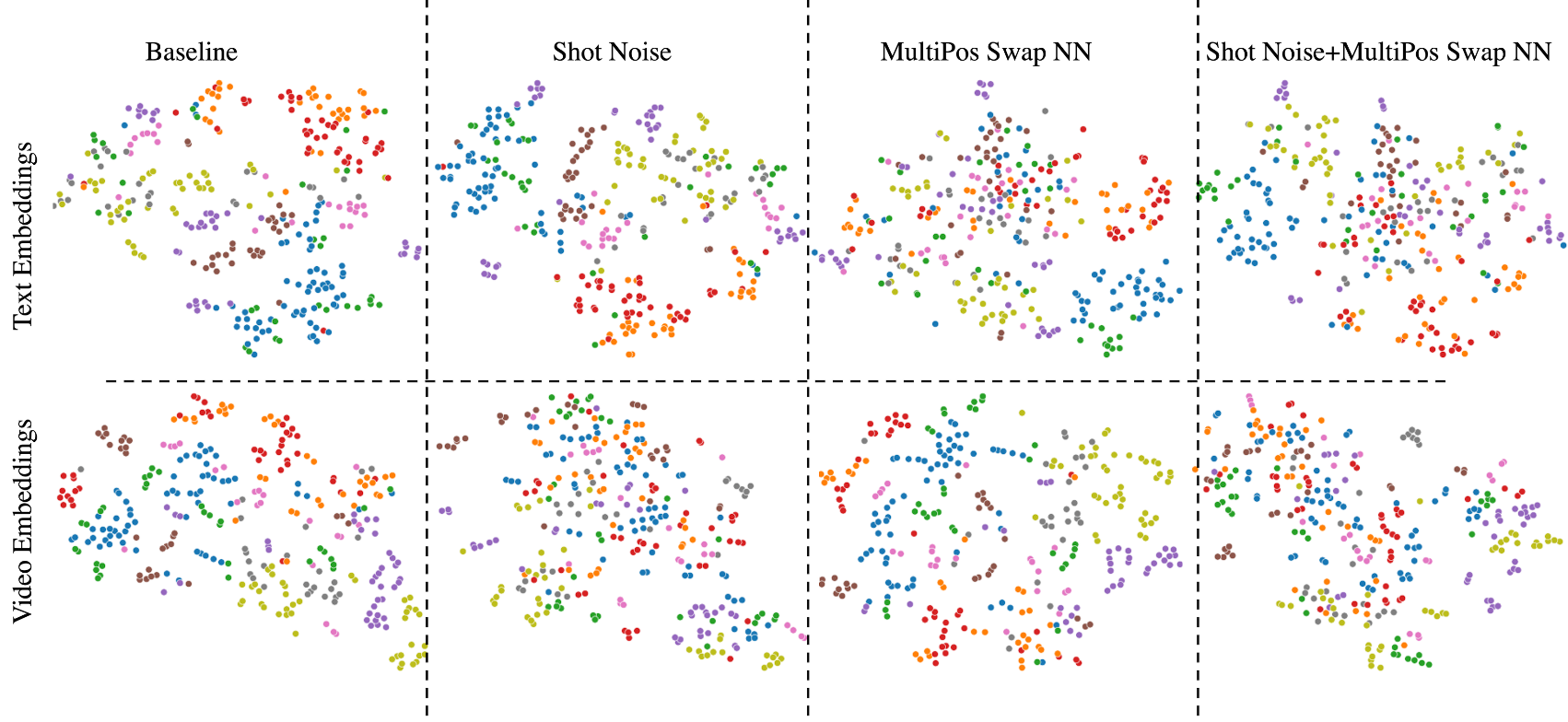}
    \caption{TSNE visualizations for output of the VideoClip model for text and video with different perturbations where colors are recipe types. This is a visualization that uses TSNE to compress the high-dimensional feature space to 2D space. Using this, we observe that when one modality is perturbed, the embedding space of the other is untouched. When both video and text are perturbed, both embedding spaces are impacted.}
    \label{fig:videoclip_tsne}
\end{figure}

\begin{figure}[t!]
    \centering
    \includegraphics[width=.95\linewidth]{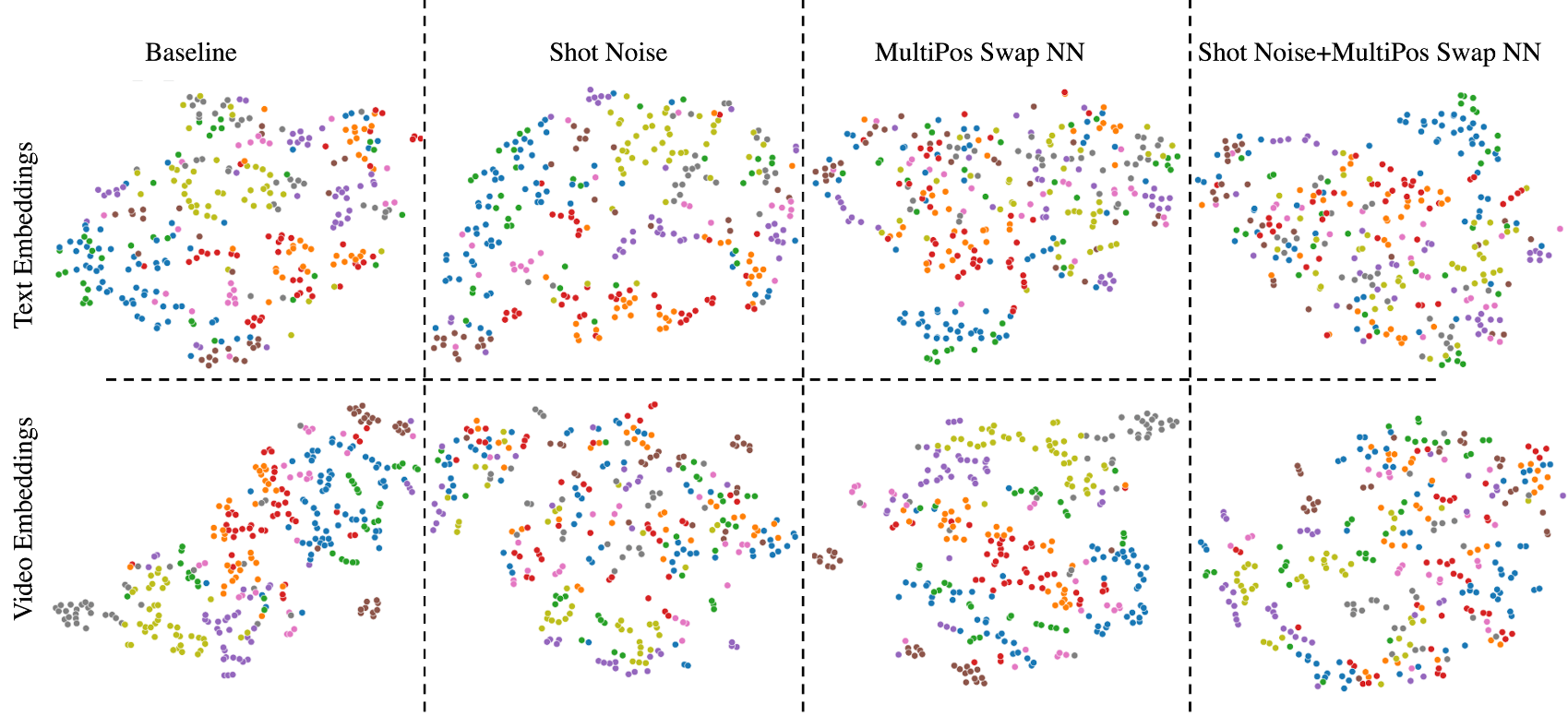}
    \caption{TSNE visualizations for output of the UniVL model for text and video with different perturbations where colors are recipe types. This is a visualization that uses TSNE to compress the high-dimensional feature space to 2D space. When one modality is perturbed, the embedding space of the other is untouched, despite cross-attention being used. When both video and text are perturbed, both embedding spaces are impacted.}
    \label{fig:univl_tsne}
\end{figure}

Figures \ref{fig:videoclip_tsne} and \ref{fig:univl_tsne} show tsne plots of the feature space for the Videoclip and UniVL models respectively when pre-trained and not fine-tuned. The colors indicate the recipe type and are just a way of visualizing space that should be more similar than videos and text from other recipes. It is important to note that the models are not trained on creating a space that clusters recipes together; therefore using recipes is just an arbitrary way of visualizing this space. In Figure \ref{fig:videoclip_tsne}, we see that when one modality is perturbed, the embedding space of the other is relatively untouched. When both video and text are perturbed, both spaces are impacted. In Figure \ref{fig:univl_tsne}, \textit{ we see a similar trend, where when one modality is perturbed, the others embedding space is relatively untouched, even though their is use of cross-attention between the two modalities.} 

\subsection{Breakdown of Perturbations}
\begin{figure}[ht!] 
    \centering
    \includegraphics[width=\linewidth]{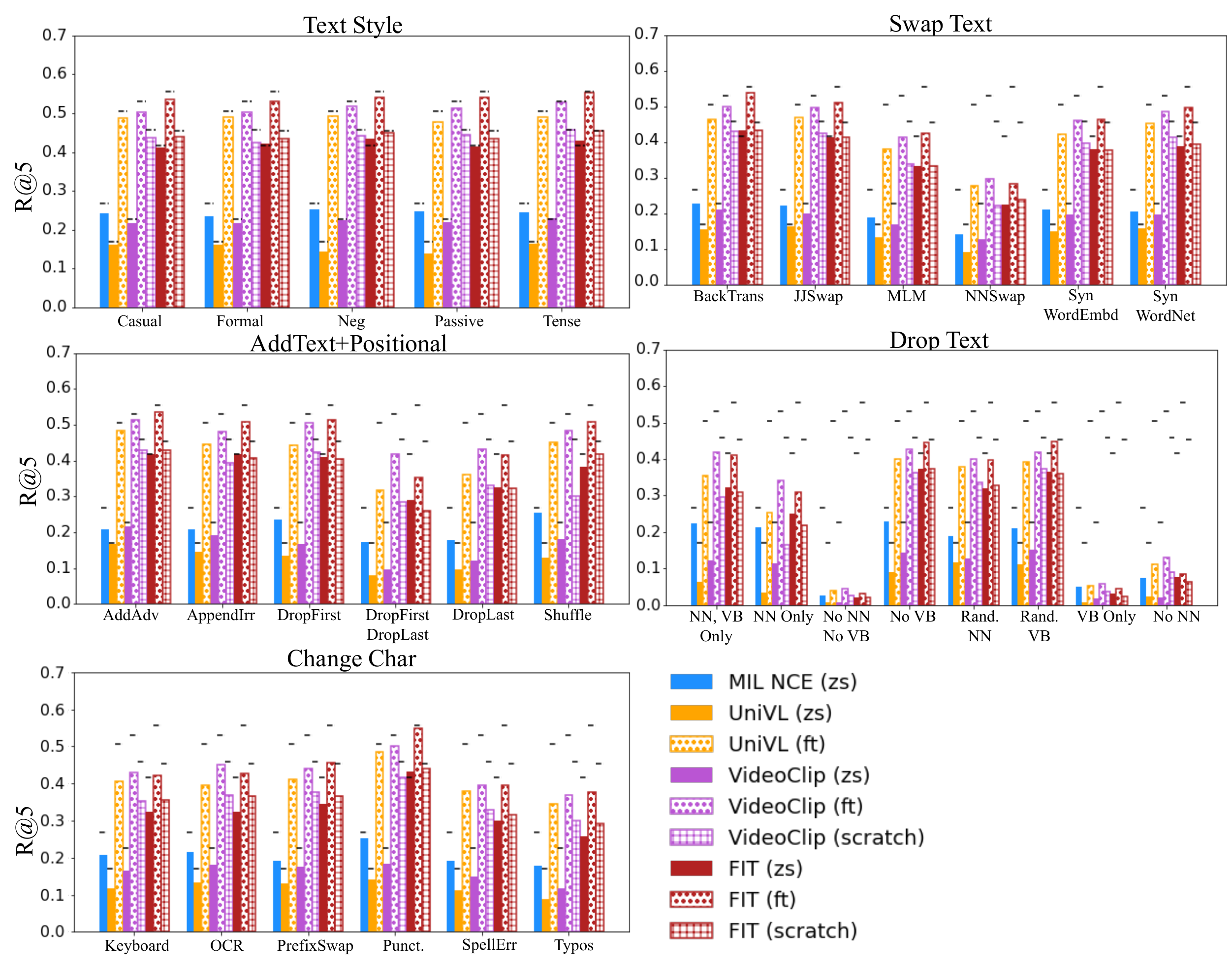}
    \caption{Perturbation specific results for text perturbations on the MSRVTT dataset. The black, horizontal lines indicate the retrieval on the clean dataset $R_c$ while the bar indicates the retrieval on the perturbed dataset $R_p$. Fine-tuned models are the highest performers but are not always more robust.}
    \label{fig:msrvtt_text_perturbations}
\end{figure}

\begin{figure}[t!] 
    \centering
    \includegraphics[width=\linewidth]{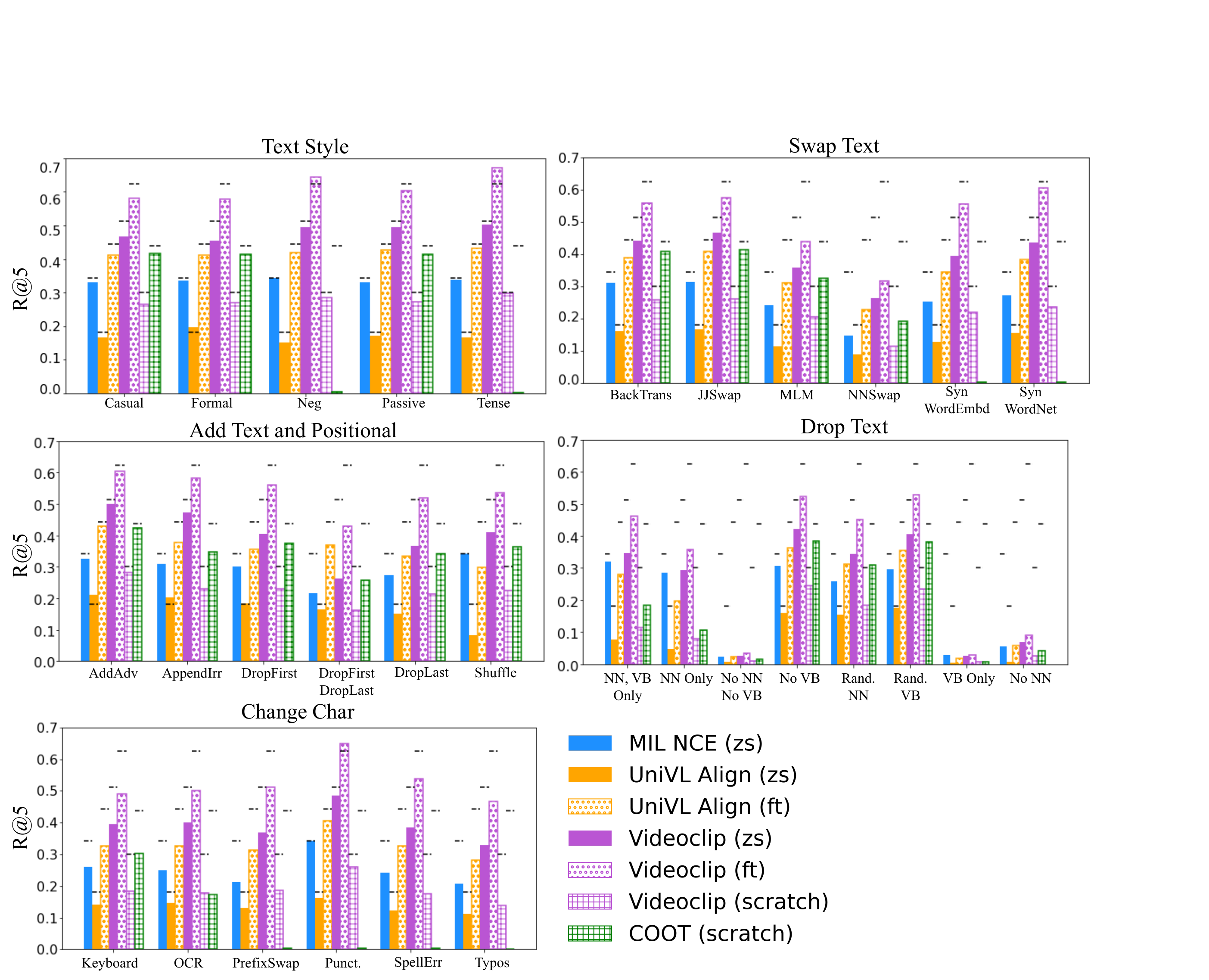}
    \caption{Perturbation specific results for text perturbations on the YouCook2 dataset. The black, horizontal lines indicate the retrieval on the clean dataset $R_c$ while the bar indicates the retrieval on the perturbed dataset $R_p$.}
    \label{fig:yc2_text_perturbations}
\end{figure}

\begin{figure}[t!]
    \centering
    \includegraphics[width=\linewidth]{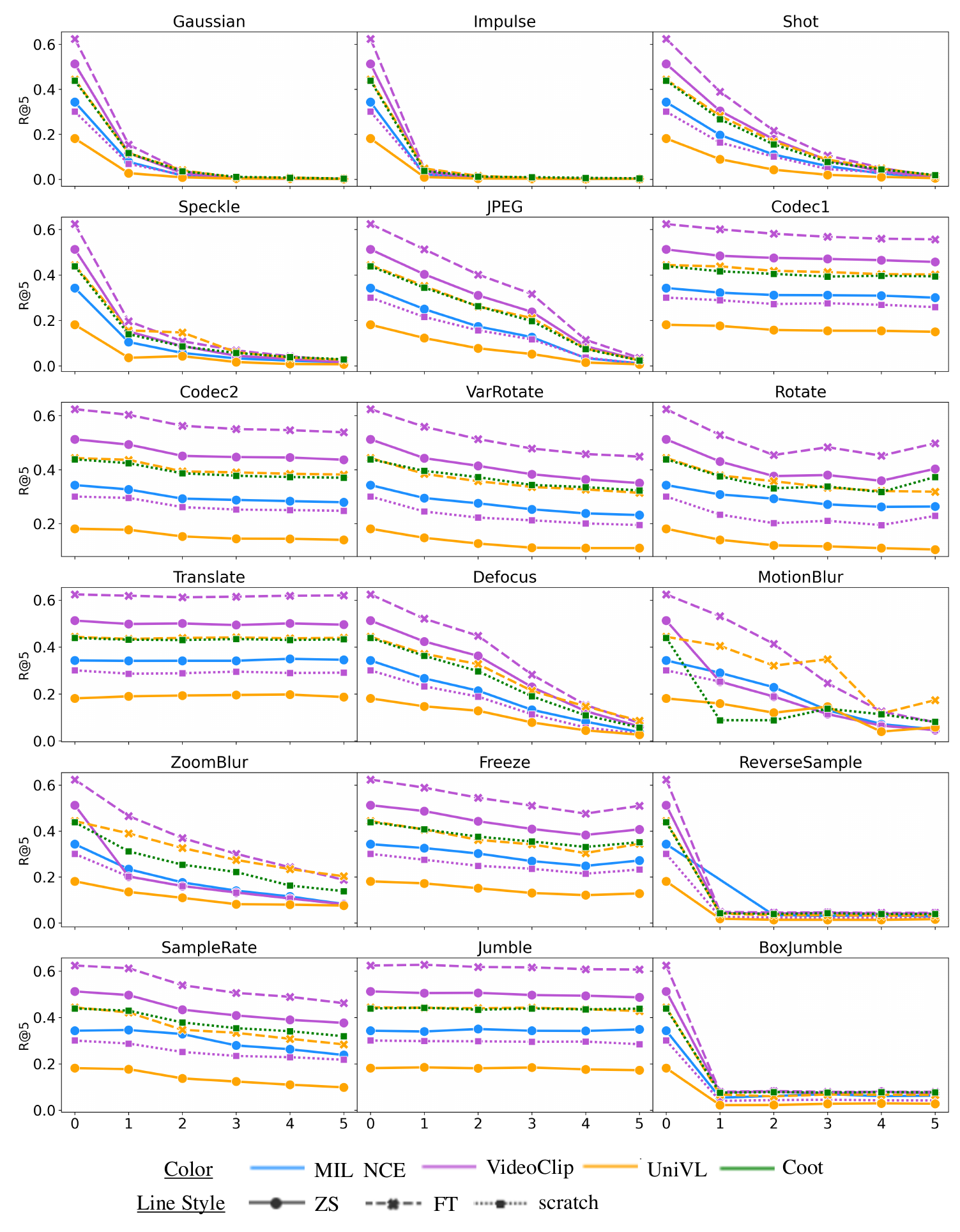}
    \caption{Performance R@5 when video is perturbed for different levels of severity on the YouCook2 dataset. Models are less robust against spacial perturbations and strongly perturbed against Temporal perturbations that maintain alignment between text and segments. When alignment is disturbed, models are no longer robust.}
    \label{fig:yc2_visual}
\end{figure}

\begin{figure}[t!]
    \centering
    \includegraphics[width=.99\linewidth]{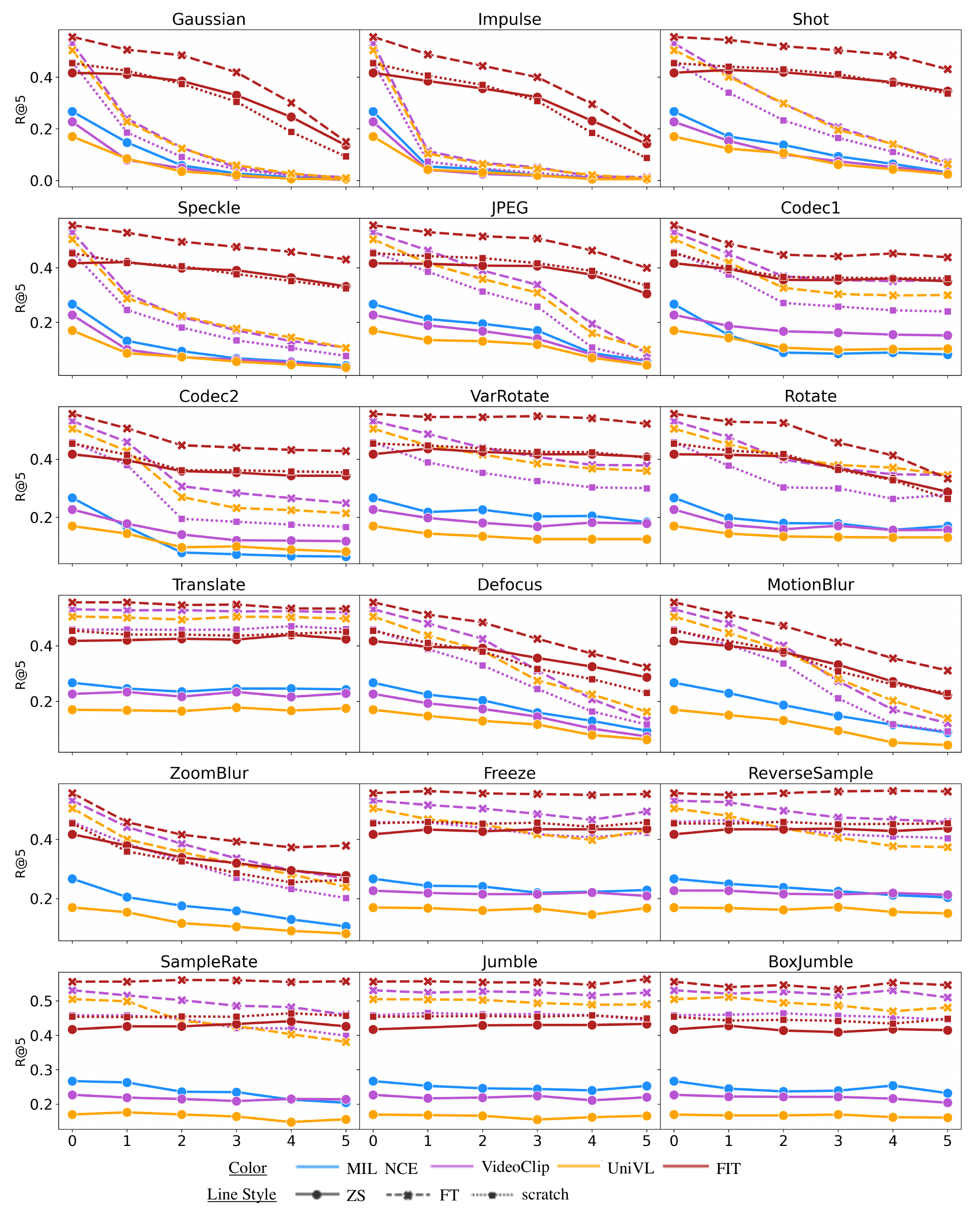}
    \caption{Performance R@5 when video is perturbed for different levels of severity on the MSRVTT dataset. Models are less robust against spacial perturbations and strongly perturbed against Temporal perturbations. Models are surprisingly robust against spatio-temporal (Digital) perturbations, struggling most with JPEG.}
    \label{fig:vtt_visual}
\end{figure}

Results for each perturbation when text is perturbed are shown in Figure \ref{fig:msrvtt_text_perturbations} and \ref{fig:yc2_text_perturbations}. In these figures, the black, dashed line indicate the original performance and the bars represent the performance when text is perturbed. The larger the difference between the top of the bar and the horizontal line indicate a larger drop in relative performance. \textit{The figures visualize the noticeable drop in performance for ChangeChar, a perturbation humans are highly robust to. It also shows how robust models are to AddText, Positional and Text Style. Finally, it shows models are surprisingly robust to the synthetic noise of Drop Text.
}
Figure \ref{fig:yc2_visual} and \ref{fig:vtt_visual} show performance R@5 over the varying levels of severity. On both datasets, the majority of models are not robust to spacial perturbations such as Noise and Blur.\textit{ FIT \cite{fit}, which uses a ViT \cite{vit, timesformer} visual encoder, is noticeably robust to spatial noise.} When text is aligned with segments, models are robust to temporal perturbations but are not on YouCook2 when text is no longer aligned with its segment from the long, complex video. However, \textit{frame order does not matter within the texts respective segment}. Models are \textit{surprisingly robust against spatio-temporal, or Digital, perturbations, struggling most with JPEG compression}.

\begin{table}[t!] 
    \centering
    \caption{Average Absolute robustness scores and their standard deviations $\gamma^a \pm \sigma$ for each category of distribution shifts for text perturbations. The UniVL model is typically the highest performer. Overall, models are very robust to text perturbations when considering the absolute score. }
    \label{tab:abs_text_perturbations}
    \resizebox{\textwidth}{!}{\begin{tabular}{{llllllll}}
    \toprule
        \textbf{MSRVTT}$\gamma^a\pm \sigma$ &                 AddText &                       Bias &                 ChangeChar &                   DropText &              Positional &                   SwapText &                  TextStyle \\
        \hline
       FIT (scratch)       &              0.96$\pm$0.02 &              0.93$\pm$0.03 &              0.90$\pm$0.05 &              0.76$\pm$0.15 &           0.90$\pm$0.07 &              0.91$\pm$0.07 &  \underline{0.99$\pm$0.01} \\
VideoClip (scratch) &              0.95$\pm$0.03 &              0.94$\pm$0.02 &              0.90$\pm$0.04 &              0.75$\pm$0.15 &           0.88$\pm$0.06 &              0.91$\pm$0.08 &              0.98$\pm$0.01 \\
FIT (zs)            &     \textbf{1.00$\pm$0.00} &              0.98$\pm$0.02 &              0.91$\pm$0.06 &              0.80$\pm$0.15 &  \textbf{0.94$\pm$0.05} &              0.95$\pm$0.08 &     \textbf{1.01$\pm$0.01} \\
MIL NCE (zs)        &              0.94$\pm$0.00 &              0.97$\pm$0.01 &  \underline{0.94$\pm$0.03} &  \underline{0.88$\pm$0.09} &  \textbf{0.94$\pm$0.04} &              0.93$\pm$0.03 &              0.98$\pm$0.01 \\
UniVL (zs)          &  \underline{0.99$\pm$0.02} &     \textbf{1.00$\pm$0.01} &     \textbf{0.95$\pm$0.02} &     \textbf{0.89$\pm$0.05} &  \textbf{0.94$\pm$0.03} &     \textbf{0.97$\pm$0.03} &              0.98$\pm$0.01 \\
VideoClip (zs)      &              0.98$\pm$0.02 &  \underline{0.99$\pm$0.01} &  \underline{0.94$\pm$0.03} &              0.86$\pm$0.06 &           0.91$\pm$0.04 &  \underline{0.96$\pm$0.03} &  \underline{0.99$\pm$0.01} \\
FIT (ft)            &              0.97$\pm$0.02 &              0.94$\pm$0.03 &              0.88$\pm$0.06 &              0.72$\pm$0.19 &           0.89$\pm$0.08 &              0.90$\pm$0.09 &              0.98$\pm$0.01 \\
UniVL (ft)          &              0.96$\pm$0.03 &              0.94$\pm$0.03 &              0.90$\pm$0.05 &              0.74$\pm$0.16 &           0.89$\pm$0.06 &              0.91$\pm$0.07 &              0.98$\pm$0.01 \\
VideoClip (ft)      &              0.97$\pm$0.02 &              0.95$\pm$0.02 &              0.90$\pm$0.05 &              0.75$\pm$0.17 &           0.93$\pm$0.04 &              0.91$\pm$0.08 &              0.98$\pm$0.01 \\
\bottomrule
        \toprule
        \textbf{YouCook2}$\gamma^a \pm \sigma$  &   AddText & Bias &                 ChangeChar &                   DropText &                 Positional &                   SwapText &               TextStyle \\

        \midrule
       COOT (scratch)      &              0.95$\pm$0.05 &  --- &              0.64$\pm$0.13 &              0.74$\pm$0.16 &              0.90$\pm$0.05 &              0.78$\pm$0.19 &           0.81$\pm$0.23 \\
VideoClip (scratch) &              0.96$\pm$0.04 &  --- &              0.89$\pm$0.04 &              0.81$\pm$0.10 &              0.91$\pm$0.03 &  \underline{0.91$\pm$0.05} &           0.98$\pm$0.01 \\

MIL NCE (zs)        &  \underline{0.97$\pm$0.01} &  --- &  \underline{0.91$\pm$0.05} &  \underline{0.85$\pm$0.14} &  \underline{0.94$\pm$0.05} &  \underline{0.91$\pm$0.06} &  \textbf{0.99$\pm$0.00} \\
UniVL (zs)          &     \textbf{1.03$\pm$0.01} &  --- &     \textbf{0.95$\pm$0.02} &     \textbf{0.90$\pm$0.07} &     \textbf{0.96$\pm$0.04} &     \textbf{0.95$\pm$0.03} &  \textbf{0.99$\pm$0.02} \\
VideoClip (zs)      &  \underline{0.97$\pm$0.02} &  --- &              0.88$\pm$0.05 &              0.73$\pm$0.17 &              0.85$\pm$0.07 &              0.88$\pm$0.07 &           0.97$\pm$0.02 \\
UniVL (ft)          &              0.96$\pm$0.04 &  --- &              0.89$\pm$0.04 &              0.76$\pm$0.15 &              0.90$\pm$0.03 &              0.90$\pm$0.07 &           0.98$\pm$0.01 \\
VideoClip (ft)      &  \underline{0.97$\pm$0.02} &  --- &              0.90$\pm$0.07 &              0.69$\pm$0.22 &              0.89$\pm$0.06 &              0.88$\pm$0.11 &  \textbf{0.99$\pm$0.04} \\
\bottomrule
        \end{tabular}}
    
\end{table}

\begin{table}[t!] 
    \centering
    \caption{Average Absolute robustness scores and their standard deviations $\gamma^a \pm \sigma$ for each category of distribution shifts for video perturbations. The UniVL model is typically the most robust model. Models are least robust to Noise, Blur and Digital.}
    \label{tab:abs_visual_robustness_scores}
    \resizebox{.80\textwidth}{!}{
            \begin{tabular}{lllllllll}
            \toprule
        \textbf{MSRVTT}$\gamma^a$ &           Blur &                     Camera &                    Digital &                      Noise &                   Temporal \\
        \hline
      FIT (scratch)       &              0.85$\pm$0.06 &              0.96$\pm$0.05 &  \underline{0.93$\pm$0.03} &  \underline{0.88$\pm$0.11} &  \underline{1.00$\pm$0.01} \\
VideoClip (scratch) &              0.79$\pm$0.11 &              0.91$\pm$0.08 &              0.78$\pm$0.10 &              0.65$\pm$0.09 &              0.98$\pm$0.02 \\
FIT (zs)            &  \underline{0.91$\pm$0.05} &     \textbf{0.99$\pm$0.04} &     \textbf{0.95$\pm$0.03} &     \textbf{0.92$\pm$0.09} &     \textbf{1.01$\pm$0.01} \\
MIL NCE (zs)        &              0.89$\pm$0.05 &              0.94$\pm$0.03 &              0.84$\pm$0.05 &              0.80$\pm$0.05 &              0.97$\pm$0.02 \\
UniVL (zs)          &     \textbf{0.93$\pm$0.04} &  \underline{0.98$\pm$0.02} &  \underline{0.93$\pm$0.03} &  \underline{0.88$\pm$0.03} &              0.99$\pm$0.01 \\
VideoClip (zs)      &  \underline{0.91$\pm$0.05} &              0.96$\pm$0.03 &              0.91$\pm$0.04 &              0.82$\pm$0.04 &              0.99$\pm$0.01 \\
FIT (ft)            &              0.86$\pm$0.06 &              0.96$\pm$0.06 &              0.91$\pm$0.04 &              0.87$\pm$0.11 &  \underline{1.00$\pm$0.01} \\
UniVL (ft)          &              0.80$\pm$0.10 &              0.92$\pm$0.06 &              0.79$\pm$0.09 &              0.63$\pm$0.11 &              0.95$\pm$0.04 \\
VideoClip (ft)      &              0.78$\pm$0.12 &              0.91$\pm$0.07 &              0.80$\pm$0.10 &              0.61$\pm$0.11 &              0.97$\pm$0.02 \\
\bottomrule
        \toprule
        \textbf{YouCook2}$\gamma^a$ &                 Blur &                     Camera &                    Digital &                      Noise &                   Temporal \\

        \hline
        COOT (scratch)      &              0.74$\pm$0.09 &              0.94$\pm$0.04 &              0.88$\pm$0.13 &              0.62$\pm$0.07 &              0.82$\pm$0.17 \\
VideoClip (scratch) &  \underline{0.83$\pm$0.07} &              0.94$\pm$0.04 &  \underline{0.91$\pm$0.09} &  \underline{0.73$\pm$0.05} &  \underline{0.87$\pm$0.12} \\

MIL NCE (zs)        &              0.81$\pm$0.08 &  \underline{0.95$\pm$0.04} &              0.90$\pm$0.10 &              0.70$\pm$0.05 &  \underline{0.87$\pm$0.13} \\
UniVL (zs)          &     \textbf{0.91$\pm$0.04} &     \textbf{0.96$\pm$0.04} &     \textbf{0.94$\pm$0.06} &     \textbf{0.84$\pm$0.02} &     \textbf{0.92$\pm$0.07} \\
VideoClip (zs)      &              0.66$\pm$0.11 &              0.91$\pm$0.06 &              0.87$\pm$0.15 &              0.54$\pm$0.08 &              0.78$\pm$0.20 \\
UniVL (ft)          &              0.82$\pm$0.10 &              0.93$\pm$0.05 &              0.89$\pm$0.13 &              0.62$\pm$0.07 &              0.80$\pm$0.17 \\
VideoClip (ft)      &              0.67$\pm$0.16 &              0.91$\pm$0.07 &              0.85$\pm$0.18 &              0.45$\pm$0.11 &              0.73$\pm$0.25 \\
\bottomrule
\end{tabular}}
\end{table}
\begin{table}[t!] 
    \centering
    \caption{Average Absolute robustness $\gamma^a$, Relative Robustness scores $\gamma^r$ and their standard deviations $ \pm \sigma$ across video, text and multimodal perturbations. FIT is noticeably more relatively robust on video perturbations likely due to the use of ViT as the visual encoder. COOT is noticeably low at text robustness, likely due to using pre-extracted text features instead of training a text encoder. }
    \label{tab:overall_robustness_by_modality}
    \resizebox{.80\textwidth}{!}{\begin{tabular}{lll|ll|ll}
    \toprule
\multirow{2}{*}{\textbf{MSRVTT}} &  \multicolumn{2}{c|}{Video} &  \multicolumn{2}{c|}{Text}  &   \multicolumn{2}{c}{Video+Text} \\
           & $\gamma^a$  &     $\gamma^r$   & $\gamma^a$      &         $\gamma^r$   &      $\gamma^a$     &        $\gamma^r$          \\
\hline
FIT (scratch)       &              0.93$\pm$0.08 &              0.84$\pm$0.18 &              0.89$\pm$0.11 &              0.75$\pm$0.25 &              0.84$\pm$0.10 &              0.65$\pm$0.22 \\
VideoClip (scratch) &              0.83$\pm$0.15 &              0.63$\pm$0.32 &              0.89$\pm$0.11 &              0.75$\pm$0.24 &              0.77$\pm$0.11 &              0.50$\pm$0.24 \\
FIT (zs)            &     \textbf{0.96$\pm$0.06} &     \textbf{0.91$\pm$0.15} &              0.92$\pm$0.11 &     \textbf{0.81$\pm$0.26} &  \underline{0.89$\pm$0.10} &     \textbf{0.73$\pm$0.24} \\
MIL NCE (zs)        &              0.89$\pm$0.08 &              0.60$\pm$0.29 &  \underline{0.94$\pm$0.05} &              0.76$\pm$0.20 &              0.87$\pm$0.06 &              0.51$\pm$0.23 \\
UniVL (zs)          &  \underline{0.94$\pm$0.05} &              0.67$\pm$0.30 &     \textbf{0.95$\pm$0.05} &              0.71$\pm$0.28 &     \textbf{0.92$\pm$0.04} &              0.54$\pm$0.24 \\
VideoClip (zs)      &              0.92$\pm$0.07 &              0.66$\pm$0.32 &  \underline{0.94$\pm$0.06} &              0.72$\pm$0.26 &              0.88$\pm$0.06 &              0.49$\pm$0.25 \\
FIT (ft)            &              0.92$\pm$0.09 &  \underline{0.86$\pm$0.15} &              0.87$\pm$0.14 &              0.77$\pm$0.24 &              0.82$\pm$0.13 &  \underline{0.67$\pm$0.23} \\
UniVL (ft)          &              0.82$\pm$0.15 &              0.65$\pm$0.29 &              0.88$\pm$0.12 &              0.77$\pm$0.23 &              0.77$\pm$0.12 &              0.54$\pm$0.23 \\
VideoClip (ft)      &              0.82$\pm$0.16 &              0.66$\pm$0.30 &              0.89$\pm$0.12 &  \underline{0.80$\pm$0.23} &              0.76$\pm$0.13 &              0.54$\pm$0.24 \\
\bottomrule

\multirow{2}{*}{\textbf{YouCook2}} &  \multicolumn{2}{c|}{Video} &  \multicolumn{2}{c|}{Text}  &   \multicolumn{2}{c}{Video+Text} \\
           & $\gamma^a$  &     $\gamma^r$   & $\gamma^a$      &         $\gamma^r$   &      $\gamma^a$     &        $\gamma^r$          \\
\hline
COOT (scratch)      &              0.79$\pm$0.16 &           0.52$\pm$0.36 &              0.77$\pm$0.17 &              0.49$\pm$0.39 &              0.68$\pm$0.11 &              0.28$\pm$0.25 \\
VideoClip (scratch) &  \underline{0.86$\pm$0.11} &           0.53$\pm$0.35 &              0.91$\pm$0.08 &              0.69$\pm$0.28 &              0.78$\pm$0.07 &              0.27$\pm$0.22 \\

MIL NCE (zs)  &              0.84$\pm$0.13 &           0.53$\pm$0.37 &  \underline{0.92$\pm$0.09} &  \underline{0.76$\pm$0.26} &  \underline{0.80$\pm$0.10} &     \textbf{0.42$\pm$0.29} \\
UniVL (zs)    &     \textbf{0.91$\pm$0.07} &           0.50$\pm$0.36 &     \textbf{0.95$\pm$0.06} &              0.73$\pm$0.31 &     \textbf{0.87$\pm$0.05} &              0.31$\pm$0.26 \\
VideoClip (zs)      &              0.74$\pm$0.19 &           0.50$\pm$0.37 &              0.86$\pm$0.13 &              0.72$\pm$0.25 &              0.65$\pm$0.13 &              0.32$\pm$0.25 \\
UniVL (ft)    &              0.80$\pm$0.16 &  \textbf{0.55$\pm$0.36} &              0.88$\pm$0.11 &              0.72$\pm$0.25 &              0.70$\pm$0.10 &              0.31$\pm$0.23 \\
VideoClip (ft)      &              0.72$\pm$0.23 &  \textbf{0.55$\pm$0.37} &              0.86$\pm$0.16 &     \textbf{0.77$\pm$0.26} &              0.58$\pm$0.16 &  \underline{0.33$\pm$0.26} \\
\bottomrule
\end{tabular}
}
\end{table}
\subsection{Absolute Robustness}
The absolute robustness scores for text perturbations are shown in Table \ref{tab:abs_text_perturbations} and visual perturbations in Table \ref{tab:abs_visual_robustness_scores}. When observing absolute robustness, UniVL Align \cite{univl},  pre-trained but not fine-tuned, is the typically the most robust model that uses a CNN visual encoder, while FIT \cite{fit} is the most robust when video is perturbed of all models. UniVL uses a cross-encoder architecture with an alignment based objective function. This differs from the relatively robust results in which it varies which model and pre-training strategy is more robust. In Table \ref{tab:abs_visual_robustness_scores}, the models struggle most with Blur, Noise and Digital.\textit{ Digital is likely challenging because it perturbs both spacial features and temporal} (see Figure \ref{fig:image_grid}). 

\subsection{Embedding Gap}
\begin{figure*}[t!]
    \centering
    \includegraphics[width=\linewidth]{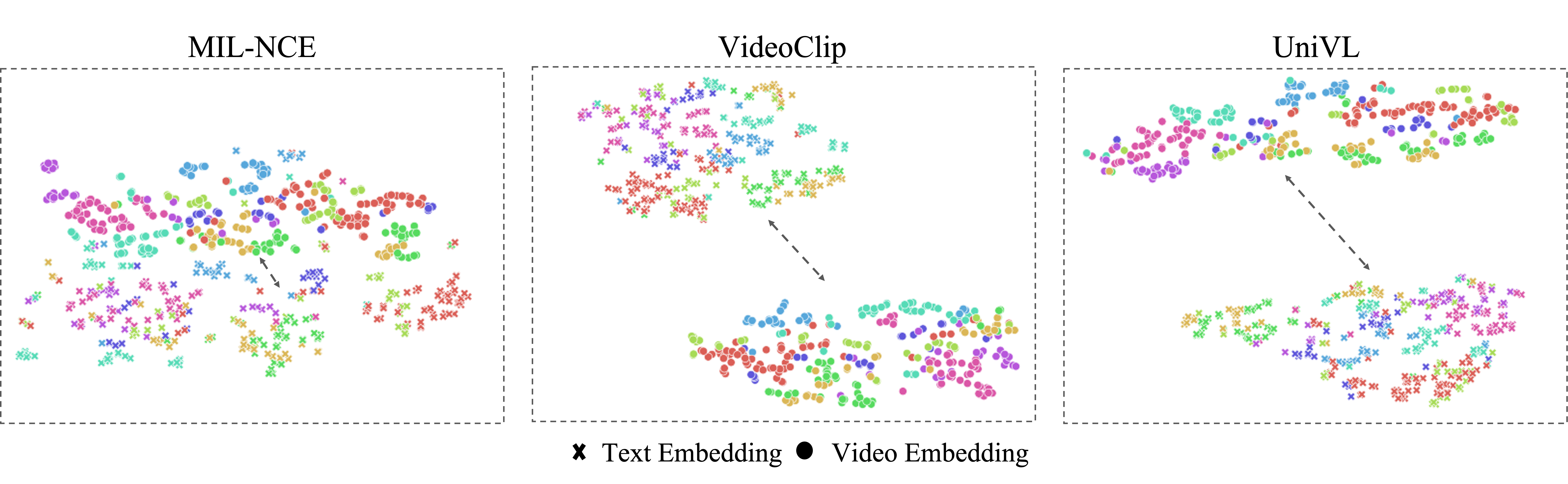}
    \caption{TNSE visualizations where video and text are in a joint space. The colors indicate the recipe of clips from the YouCook2 dataset. The shapes of each point indicate either text or video embeddings. 
    }
    \label{fig:embedding_gap}
\end{figure*}

Figure \ref{fig:embedding_gap} shows a TSNE \cite{tsne} plot where we use recipe type from YouCook2 as color. Text embeddings are indicated by $X$ and visual embeddings are indicated by a circle.  When visualized, even though the VideoClip \cite{videoclip} and UniVL \cite{univl} models significantly outperform the MIL-NCE model \cite{miech2020end} on downstream tasks, the text and visual embeddings show a large gap in the latent space compared to MIL-NCE. Future work should look at why these higher performing models produce multimodal embeddings that look distant from each other.

\subsection{Changing of Semantic Meaning} 
Some of the perturbations result in semantically incorrect captions when compared to the original. An example of this is under \textit{Text Style}, the perturbation \textit{Neg}, where ``a girl who does gymnastics'' becomes ``a girl who does not gymnastics''. Figure \ref{fig:yc2_text_perturbations} and \ref{fig:msrvtt_text_perturbations} shows the results for YouCook2-P and MSRVTT-P respectfully. We observe that model text-to-video retrieval drop in performance is noticeably low when using the negative caption. The exception is with COOT on the YouCook2 dataset, (Figure \ref{fig:yc2_text_perturbations}) which uses pre-extracted text features without fine-tuning the modality-specific text encoder. Another example where we know that the statement is now incorrect is with \textit{Gender Swap}, where a female reference is made male and vice-versa. Figure \ref{fig:bias_perts} shows performance on this perturbation and while relative model performance decreases, this drop is low. We observe similar behavior on perturbations that do not guarantee a change in semantic meaning but are likely to, including \textit{SwapText} perturbations \textit{JJSwap} and \textit{MLM}. However, when we swap a word with a new noun in \textit{NNSwap} under \textit{SwapText}, we find that models' retrieval scores drop more noticeably. Additionally, models are highly robust when words in the text are shuffled under \textit{Positional} perturbations. Shuffling the word order can change the adjectives assigned to nouns as well as the adverbs and verbs, potentially altering the semantic meaning of the text. This indicates that \textit{models may be attending most to objects rather than the overall semantic meaning as defined by word order for adjectives, verbs, and adverbs.}

\subsection{Perturb Modality}
Table \ref{tab:overall_robustness_by_modality} shows the relative and absolute robustness across the video, text or both being perturbed for each dataset. FIT \cite{fit} is noticeably more relatively robust on video perturbations, likely due to the use of ViT as the visual encoder. COOT \cite{COOT} is noticeably low at text robustness, likely due to using pre-extracted text features instead of training a text encoder. When both is perturbed, UniVL has the highest absolute robustness for both datasets. UniVL uses cross-attention during training as opposed to a two-encoder architecture like VideoClip. This could be a potential reason for the high absolute robustness UniVL demonstrates when using zero-shot evaluation.

\subsection{MRSVTT QA}
To understand if these findings transcend to other video-language tasks, we evaluated VideoClip on the multiple choice VideoQA task with results in Table \ref{tab:vttqa_text} and \ref{tab:vttqa_video}.
%
When only text is perturbed, the the difference between pre-training strategy is not consistent unlike the text-to-video retrieval task. Unlike the previous task, zero-shot typically is the least robust and scratch is as robust or more than fine-tuning. This indicates that when the task is between smaller candidates, pre-training on a large corpus of data may be less necessary for both performance and robustness. 

When text is perturbed, the zero-shot model is typically more robust. Very different findings between the two modalities. This is likely because this task is similar to video-to-text retrieval and we have already observed models are less robust to visual perturbations. Zero-shot therefore is more relatively robust because of the nature of the pre-training dataset of HowTo100m \cite{howto100m} being a variety of noisy YouTube videos.

\begin{table}[t!] 
    \centering
    \caption{Average Relative robustness scores and their standard deviations $\gamma^r \pm \sigma$ for text categories on the MSRVTT-QA for the Videoclip model and its training variations.}
    \label{tab:vttqa_text}
    \resizebox{.80\textwidth}{!}{\begin{tabular}{llllllll}
\toprule
VideoQA &                 AddText &                    Bias &              ChangeChar &                DropText &              Positional &                SwapText &             TextStyle \\
\midrule
scratch &  \textbf{0.99$\pm$0.01} &           0.98$\pm$0.02 &  \textbf{0.96$\pm$0.02} &  \textbf{0.77$\pm$0.23} &  \textbf{0.97$\pm$0.03} &           0.94$\pm$0.07 &  \textbf{1.0$\pm$0.0} \\
zeroshot      &           0.97$\pm$0.03 &           0.98$\pm$0.01 &           0.88$\pm$0.07 &           0.67$\pm$0.26 &            0.9$\pm$0.06 &           0.89$\pm$0.09 &         0.98$\pm$0.01 \\
finetune      &  \textbf{0.99$\pm$0.01} &  \textbf{0.99$\pm$0.01} &  \textbf{0.96$\pm$0.02} &           0.75$\pm$0.25 &           0.96$\pm$0.03 &  \textbf{0.95$\pm$0.06} &         0.99$\pm$0.01 \\
\bottomrule\end{tabular}}
\end{table}

\begin{table}[t!] 
    \centering
    \caption{Average Relative robustness scores and their standard deviations $\gamma^r \pm \sigma$ for visual categories on the MSRVTT-QA for the Videoclip model and its training variations.}
    \label{tab:vttqa_video}
    \resizebox{.65\textwidth}{!}{\begin{tabular}{llllll}
\toprule
VideoQA &                   Blur &                  Camera &                Digital &                   Noise &               Temporal \\

\midrule
scratch &          0.75$\pm$0.14 &           0.89$\pm$0.08 &          0.71$\pm$0.16 &            0.46$\pm$0.2 &          0.93$\pm$0.03 \\
zeroshot      &  \textbf{0.8$\pm$0.15} &  \textbf{0.96$\pm$0.07} &  \textbf{0.8$\pm$0.12} &  \textbf{0.51$\pm$0.18} &  \textbf{1.0$\pm$0.01} \\
finetune      &          0.78$\pm$0.14 &           0.91$\pm$0.08 &          0.79$\pm$0.14 &            0.49$\pm$0.2 &          0.95$\pm$0.02 \\
\bottomrule
\end{tabular}}
    \end{table}


\begin{table}[t!] 
    \centering
    \caption{Distribution Shift evaluation using average Relative robustness scores and their standard deviations $\gamma^r \pm \sigma$ and Average Absolute robustness scores and their standard deviations $\gamma^a \pm \sigma$ on MSRVTT and YouCook2 captions respectively when over Natural vs. Machine vs. Artificial (Positional and DropText). }
    \label{tab:msrvtt_subtype}
    \resizebox{.80\textwidth}{!}{\begin{tabular}{lr|r|r|r|r|r}
        \toprule
        \multirow{2}{*}{\textbf{MSRVTT}}   &  \multicolumn{2}{c}{Natural} &  \multicolumn{2}{c}{Machine} &  \multicolumn{2}{c}{Synthetic} \\
           &     $\gamma^a$ & $\gamma^r$  & $\gamma^a$ & $\gamma^r$ & $\gamma^a$ & $\gamma^r$ \\
        \hline
        FIT (scratch)       &              0.90$\pm$0.09 &              0.78$\pm$0.20 &           0.94$\pm$0.05 &              0.87$\pm$0.11 &              0.82$\pm$0.14 &              0.61$\pm$0.31 \\
VideoClip (scratch) &              0.90$\pm$0.10 &              0.77$\pm$0.21 &           0.94$\pm$0.04 &              0.87$\pm$0.10 &              0.82$\pm$0.14 &              0.60$\pm$0.30 \\
FIT (zs)            &              0.93$\pm$0.10 &     \textbf{0.84$\pm$0.23} &  \textbf{0.97$\pm$0.05} &     \textbf{0.92$\pm$0.13} &              0.86$\pm$0.14 &  \underline{0.67$\pm$0.33} \\
MIL NCE (zs)        &              0.93$\pm$0.04 &              0.73$\pm$0.15 &           0.96$\pm$0.02 &              0.85$\pm$0.09 &  \underline{0.91$\pm$0.07} &              0.66$\pm$0.27 \\
UniVL (zs)          &     \textbf{0.97$\pm$0.03} &  \underline{0.80$\pm$0.18} &  \textbf{0.97$\pm$0.02} &              0.85$\pm$0.14 &     \textbf{0.92$\pm$0.05} &              0.50$\pm$0.30 \\
VideoClip (zs)      &  \underline{0.95$\pm$0.04} &              0.78$\pm$0.16 &  \textbf{0.97$\pm$0.03} &              0.86$\pm$0.14 &              0.89$\pm$0.06 &              0.52$\pm$0.28 \\
FIT (ft)            &              0.88$\pm$0.11 &              0.79$\pm$0.21 &           0.93$\pm$0.06 &              0.88$\pm$0.10 &              0.80$\pm$0.17 &              0.63$\pm$0.31 \\
UniVL (ft)          &              0.89$\pm$0.09 &              0.79$\pm$0.18 &           0.94$\pm$0.05 &              0.88$\pm$0.09 &              0.81$\pm$0.15 &              0.63$\pm$0.29 \\
VideoClip (ft)      &              0.90$\pm$0.10 &  \underline{0.80$\pm$0.18} &           0.94$\pm$0.05 &  \underline{0.89$\pm$0.09} &              0.83$\pm$0.16 &     \textbf{0.68$\pm$0.30} \\
        \bottomrule
        \toprule
        \textbf{YouCook2}    &           $\gamma^a$ & $\gamma^r$  & $\gamma^a$ & $\gamma^r$ & $\gamma^a$ & $\gamma^r$      \\
        \hline
        COOT (scratch)      &              0.90$\pm$0.10 &  \textbf{0.76$\pm$0.24} &              0.75$\pm$0.19 &              0.44$\pm$0.44 &              0.76$\pm$0.16 &              0.45$\pm$0.37 \\

VideoClip (scratch) &  \underline{0.91$\pm$0.07} &           0.70$\pm$0.23 &  \underline{0.95$\pm$0.05} &              0.83$\pm$0.18 &              0.86$\pm$0.09 &              0.52$\pm$0.30 \\

MIL NCE (zs)        &  \underline{0.91$\pm$0.08} &           0.74$\pm$0.23 &  \underline{0.95$\pm$0.05} &              0.86$\pm$0.15 &  \underline{0.89$\pm$0.11} &     \textbf{0.67$\pm$0.32} \\
UniVL (zs)          &     \textbf{0.95$\pm$0.04} &           0.73$\pm$0.21 &     \textbf{0.98$\pm$0.03} &  \underline{0.88$\pm$0.17} &     \textbf{0.93$\pm$0.07} &              0.59$\pm$0.37 \\
VideoClip (zs)      &              0.87$\pm$0.09 &           0.75$\pm$0.18 &              0.93$\pm$0.06 &              0.86$\pm$0.11 &              0.79$\pm$0.15 &              0.59$\pm$0.29 \\

UniVL (ft)          &              0.89$\pm$0.08 &           0.75$\pm$0.19 &              0.93$\pm$0.05 &              0.85$\pm$0.12 &              0.82$\pm$0.13 &              0.59$\pm$0.30 \\
VideoClip (ft)      &              0.85$\pm$0.12 &  \textbf{0.76$\pm$0.19} &  \underline{0.95$\pm$0.06} &     \textbf{0.91$\pm$0.10} &              0.78$\pm$0.20 &  \underline{0.65$\pm$0.32} \\
        \bottomrule
\end{tabular}}

\end{table}

\subsection{Natural vs. Machine vs. Artificial Text Perturbations}
To understand models based on the sub-categories of natural, machine-learning based and synthetic, we aggregate scores these categories in Table \ref{tab:msrvtt_subtype} (see Figure \ref{fig:video_perturbations} for more details on these categories). Synthetic perturbations are those in DropText and Positional as they would be rare occurrences in the real world, although still possible with automatic speech recognition for example. When only text is perturbed, UniVL on zero-shot learning is typically more relatively robust for all three categories, although close to the MIL NCE robustness on synthetic perturbations. 


\end{document}